\documentclass[journal,comsoc]{IEEEtran}

\usepackage[T1]{fontenc}

\usepackage{graphicx}
\usepackage[cmintegrals]{newtxmath}
\usepackage{booktabs}
\usepackage{textcomp}
\usepackage{linguex}
\usepackage[linguistics,edges]{forest}
\usepackage[shortlabels]{enumitem}
\usepackage{tikz}
\usepackage{tikz-qtree}
\usetikzlibrary{trees} 
\usepackage{array}
\newcommand{\readded}

\usepackage{xcolor}
\usepackage{booktabs}
\usepackage{multirow}
\usepackage{metalogo}
\usepackage{smartdiagram}
\usepackage{algorithmic}
\usepackage{tabularx,multirow,array}
\usepackage{url}
\usepackage{csquotes} 
\usepackage{arydshln} 
\usepackage{color, xcolor, soul}
\usepackage{makecell}
\usepackage{cite}
\usepackage{etoolbox}
\usepackage{threeparttable}
\usepackage{pifont}
\usepackage{array}
\newcolumntype{L}[1]{>{\raggedright\let\newline\\\arraybackslash\hspace{0pt}}m{#1}}
\newcolumntype{C}[1]{>{\centering\let\newline\\\arraybackslash\hspace{0pt}}m{#1}}

\RequirePackage{tikz}
\RequirePackage{etoolbox}
\RequirePackage{xparse}
\RequirePackage{xstring}
\usetikzlibrary{backgrounds,
	calc,
	fadings,
	shadows,
	shapes.arrows,
	shapes.symbols
}
\edef\RestoreEndlinechar{%
	\endlinechar=\the\endlinechar\relax
}
\usepackage{smartdiagram}
\RestoreEndlinechar

\newcommand{\cmark}{\ding{51}}%
\newcolumntype{P}[1]{>{\centering\arraybackslash}p{#1}}
\usepackage{acronym}
\usepackage{balance}

\begin{document}

\title{Machine Learning for Security in Vehicular Networks: A Comprehensive Survey}

\author{Anum~Talpur,~\IEEEmembership{Member,~IEEE,}
	and~Mohan~Gurusamy,~\IEEEmembership{Senior~Member,~IEEE}
	\thanks{A. Talpur and M. Gurusamy are with the Department of Electrical and Computer Engineering, National University of Singapore, Singapore (email: anum.talpur@u.nus.edu; gmohan@nus.edu.sg).}}%
\maketitle

\begin{abstract}
Machine Learning (ML) has emerged as an attractive and viable technique to provide effective solutions for a wide range of application domains. An important application domain is vehicular networks wherein ML-based approaches are found to be very useful to address various problems. The use of wireless communication between vehicular nodes and/or infrastructure makes it vulnerable to different types of attacks. In this regard, ML and its variants are gaining popularity to detect attacks and deal with different kinds of security issues in vehicular communication. In this paper, we present a comprehensive survey of ML-based techniques for different security issues in vehicular networks. We first briefly introduce the basics of vehicular networks and different types of communications. Apart from the traditional vehicular networks, we also consider modern vehicular network architectures. We propose a taxonomy of security attacks in vehicular networks and discuss various security challenges and requirements. We classify the ML techniques developed in the literature according to their use in vehicular network applications. We explain the solution approaches and working principles of these ML techniques in addressing various security challenges and provide insightful discussion. The limitations and challenges in using ML-based methods in vehicular networks are discussed. Finally, we present observations and lessons learned before we conclude our work.
\end{abstract}


\IEEEpeerreviewmaketitle

\section{Introduction}
\IEEEPARstart{R}{ecently}, there has been an increasing interest in vehicular network architectures, protocols, and applications \cite{TransSurvey, IntelligentTransSurvey, VANETSurvey2, VANETSurvey3}. A vehicular network is an ad-hoc network used to assist the transportation system in multiple applications such as road safety, traffic management, speed controlling, infotainment services on vehicles, assistance to driverless cars, and so on. To support these applications, several variants of vehicular networks have emerged which are driven by modern technologies such as fifth-generation (5G), Internet of Things (IoT), Software-Defined Networking (SDN), edge-computing, and cloud-computing. The integration of advanced technologies demands more intelligent solutions to address the challenges arising from the diverse nature of vehicular designs. The automotive industry realizes the need for new protocols and techniques to be compatible with the new network trends and variants. \par
The traditional architecture of vehicular ad-hoc networks (VANET) used to assist autonomous and non-autonomous vehicles is comprised of on-board unit (OBU), edge devices, road side unit (RSU), centralized controllers and trusted authority (TA). A vehicular network communicates to an edge network which in turn is connected to a backbone network through a wired or wireless medium. Data transfer takes place between vehicles and different levels of networks resulting in different types of communication \cite{InterVehicleComm}, i.e. vehicle-to-vehicle (V2V), vehicle-to-RSU (V2R), infrastructure-to-infrastructure (I2I), vehicle-to-infrastructure (V2I) and many more. The improved connectivity and increased number of communication channels and access points have led to several breakthroughs. At the same time, they pose several challenges that need to be considered while designing vehicular solutions, among which the important ones are security and privacy of data \cite{ref42}. \par 

Vehicular networks are vulnerable to different types of attacks \cite{AttacksSurvey}. A number of cryptographic solutions have been proposed in the past to deal with different types of security issues \cite{Ref3}. Some commonly used traditional authentication techniques such as password protection, key-based authentication, and biometric security techniques can also be used to authenticate cars in vehicular networks. However, such techniques fail to validate whether the transferred value is spoofed or real. In addition, these techniques are difficult to achieve high accuracy and implement in low-powered vehicle security systems. Recently, there has been an immense interest in using machine learning (ML) to deal with vehicle security to get faster and highly-accurate attack predictions. ML is one of the most promising technologies in today's world of wireless networks \cite{ML_Wireless, ML_Wireless3, ML_Wireless4}. A wide range of ML techniques are used in the literature for different wireless applications \cite{ML_Wireless2}. In this article, we carry out a comprehensive survey to bring out different security challenges and requirements for vehicular networks and present an in-depth study of the state-of-the-art works that use ML algorithms to solve various security problems. The review papers in the current literature either consider the security problem in vehicular networks or ML-based applications in vehicular networks, as shown in Table. \ref{tab:surveyComp}. Different from such survey works, we present a comprehensive survey focusing exclusively on ML-based security solutions in vehicular networks.

\subsection{Related Surveys}

\begin{table*}[htbp]
  \centering
  \caption{Comparison of Surveys on ML and/or Security in Vehicular Networks}
    \begin{tabular}{C{1cm}lL{5.3cm}C{3.6cm}P{2.5cm}P{2cm}}
    \toprule
    \multicolumn{1}{c}{Citation} & \multicolumn{1}{c}{Year} & \multicolumn{1}{c}{Title} & Focused Area & Type of Network & ML context \\
    \midrule
    \cite{ref83} & 2015  & A Security and Privacy Review of VANETs & Security and Privacy & VANET & No \\
    \cite{ref84} & 2016  & Trust Management for Vehicular Networks: An Adversary-Oriented Overview & Trust Management & VANET & No \\
    \cite{VANETAttack} & 2017  & A Survey of Attacks and Detection Mechanisms on Intelligent Transportation Systems: VANETs and IoV & Attacks & VANET and IoV & Traditional ML only \\
    \cite{ref82} & 2018  & Security and Privacy in Location-Based Services for Vehicular and Mobile Communications: An Overview, Challenges, and Countermeasures & Security and Privacy in LBS & VANET & No \\
    \cite{ref86} & 2019  & Toward Intelligent Vehicular Networks: A Machine Learning Framework & Decision Making & VANET, IoV and 5GVN & ML and DL \\
    \cite{Ref3}  & 2019  & A Survey of Security Services, Attacks, and Applications for Vehicular Ad Hoc Networks (VANETs) & Security and Privacy & VANET & No \\
    \cite{ref42} & 2019  & A Survey on Recent Advances in Vehicular Network Security, Trust, and Privacy & Security, Trust and Privacy & VANET & No \\
    \cite{VANETSurvey} & 2019  & A Comprehensive Survey on VANET Security Services in Traffic Management System & Security and Privacy & VANET & No \\
    \cite{NS_IoV} & 2019  & A Survey on Internet of Vehicles: Applications, Security Issues \& Solutions & Security Attacks and Other Application & VANET and IoV & Limited  \\
    \cite{NS_VANET} & 2019  & A Survey on Security Attacks in VANETs: Communication, Applications and Challenges & Security and Privacy in VANET Applications & VANET & Limited \\
   \cite{RL2} & 2019  & Applications of Deep Reinforcement Learning in Communications and Networking: A Survey & Control, Caching, Offloading, Security, Connectivity, Routing, Scheduling, and Data Collection  & Communication Network & DRL \\
   \cite{NS_AIV2X} & 2019 & Artificial Intelligence for Vehicle-to-Everything: A Survey & Safety, Congestion, Demand and Supply Applications, Navigations, Security and Vehicle Platoons & VANET and IoV & ML and DL \\
   \cite{DLits} & 2019  & Deep Learning for Intelligent Transportation Systems: A Survey of Emerging Trends & Prediction, Control and Optimization related Transportation Applications & ITS   & DL \\
    \cite{nRef9SURVEY} & 2020  & Comprehensive Survey of Machine Learning Approaches in Cognitive Radio-Based Vehicular Ad Hoc Networks & Cognitive Radio-based Vehicular Applications & Vehicular Networks and all of its Variants & ML and DL \\
    \cite{NS_DRLTrans} & 2020  & Deep Reinforcement Learning for Intelligent Transportation Systems: A Survey & Traffic Signal Control, Autonomous Driving, Energy Management, Road Control and other Applications & ITS   & DRL \\
    \cite{NS_SPT_MIoT} & 2020  & Security, Privacy and Trust for Smart Mobile-Internet of Things (M-IoT): A Survey & Security, Trust and Privacy in Mobile-IoT Applications & MIoT  & Limited \\
    \cite{NS_IoTSec} & 2020 & A Survey on the Internet of Things (IoT) Forensics: Challenges, Approaches, and Open Issues & Security Attacks & IoT & Limited \\
    \cite{NS_MLIoTSec1} & 2020 & A Survey of Machine and Deep Learning Methods for Internet of Things (IoT) Security & Security Threats Types and Threats Surface & IoT & ML and DL \\
    \cite{NS_MLIoTSec2} & 2020 & Machine Learning in IoT Security: Current Solutions and Future Challenges & Authentication and Security Attacks & IoT & ML and DL \\
    \cite{NS_RecentV2X} & 2020  & Recent Advances and Challenges in Security and Privacy for V2X Communications & Security, Trust and Privacy & Vehicular Networks and all of its Variants & Limited \\
    \cite{FLPS2} & 2020 & Federated Learning for Vehicular Internet of Things: Recent Advances and Open Issues & Perception, Networking, Computing and Security & Vehicular-IoT & FL \\
    \cite{AD1} & 2021  & A Survey of Deep Learning Applications to Autonomous Vehicle Control & Lateral and Longitudinal Vehicle Control System & Autonomous Vehicle Network & ML and DL \\
    Our Work & 2021  & Machine Learning for Security in Vehicular Networks: A Comprehensive Survey & Security, Trust and Privacy & Vehicular Networks and all of its Variants & Wide range of ML and other learning types \\
    \bottomrule
    \end{tabular}%
  \label{tab:surveyComp}%
\end{table*}%
The use of ML in vehicular networks is gaining significant attention to address multiple challenges \cite{ref86}. Several survey works exist in the literature which cover different security problems in vehicular networks and discuss challenges with solutions proposed in the literature. A systematic comparison of available survey works with our survey work in vehicular networks is given in Table. \ref{tab:surveyComp}. \par 
In \cite{ref83}, a survey is carried out on security and privacy in VANET. This work states security challenges and discusses available solutions for security, authentication, and privacy, but doesn't present the use of ML in providing security solutions. It focuses mostly on using digital signature algorithms for VANET security. \par 
Sakiz et al. \cite{VANETAttack} survey the literature related to attacks in VANETs and Internet of Vehicles (IoVs). This work mentions the use of ML to solve different attack problems. It briefly summarizes the attack prevention mechanisms which use cryptography and ML in their framework. Only the traditional supervised and unsupervised ML approaches are discussed. Also, this work does not discuss privacy and trust issues in vehicular networks. A trust management-focused work is proposed in \cite{ref84} for vehicular networks. This work presents an adversary-oriented overview of trust management techniques. The authors classify the security frameworks into two categories; cryptography and trust. This survey differentiates and explains possible relations of trust and cryptography. Further, it briefly explains trust-based solutions with a tabular comparison of existing techniques. None of these techniques considers the use of ML techniques. It concludes that existing trust mechanisms are not enough in multiple VANET scenarios and it calls for future work on more intelligent mechanisms.\par
Asuquo et al. \cite{ref82} present a review on the security and privacy of location-based services (LBS) in vehicular networks. This work elaborates on open challenges on location privacy in vehicular networks. It classifies the work in the literature on location privacy into two groups, cryptographic mechanisms (like signature, key-based cryptography, hashing and so on) and privacy-enhancing schemes (like mixed zones, obfuscation, silent period and so on). It further explains these two groups comprehensively with a tabular comparison. In addition, it also mentions the drawbacks of such schemes in different vehicular scenarios. It does not consider any ML-based solutions.\par
Recently, researchers have started to explore more into vehicular networks and their applications. Liang et al. \cite{ref86} discuss briefly the use of ML frameworks in providing solutions to address the challenges of high-mobility vehicular networks and their variants including, IoV and 5G-based vehicular networks (5GVN). The dynamics of high-mobility networks covered in \cite{ref86} include network topology, channel estimation, traffic prediction, trajectory prediction, congestion control, and so on. It also outlines the use of ML for intrusion detection in connected vehicles. Most of the works presented in this survey are about network optimization in high mobility scenarios. Sheikh et. al \cite{Ref3} perform a survey on the security problems in VANET. It details the basics of VANET architecture and its security challenges in a comprehensive way. It briefs the state-of-the-art methods used for security and authentication problems. This work focuses on symmetric cryptography, asymmetric cryptography, identity-based cryptography, and signature schemes. However, the scope of this survey does not cover ML for security problems. Lu et. al \cite{ref42} study recent advances on security, privacy, and trust management in VANET. This survey starts with a brief background of VANET architecture and related security challenges. It discusses different attacks over security services and available cryptographic methods. Apart from security services, it elaborates on the challenges that VANET face related to location privacy. It also describes the significance of trust management in VANET and elaborates trust models available in the literature to accomplish this task. This work does not cover the importance or use of ML algorithms for securing VANET. A similar kind of work is done in \cite{VANETSurvey} where authors describe the state-of-the-art methods used to secure and preserve privacy in a VANET architecture. This work also presents a classification taxonomy for attacks and authentication mechanisms in VANETs with a major focus on authentication mechanisms. \par 
An in-depth survey of IoV is presented in \cite{NS_IoV}, where authors compare it with VANET. The authors categorize IoV applications into four categories including, safety, comfort and infotainment, traffic efficiency, and health care, and briefly discuss their usage in driving coordination and emergency warning. An elaborative discussion is made on attacks over IoV networks. But the context of ML is minimally explored in this work. Another application-based survey to discuss vehicular attacks is carried out in \cite{NS_VANET}. Similar to \cite{NS_IoV}, this work also explores the area of ML and other learning types only a little.\par 
A deep reinforcement learning (DRL)-focused survey is presented in \cite{RL2}. The scope of this survey is not specific to vehicular networks. In addition, this survey briefly covers network control, caching, offloading, routing, scheduling, and connectivity applications along with one section on DRL usage in maintaining network security. Later, Tong et. al \cite{NS_AIV2X} carried out an ML-specific survey for vehicular communication networks. It explores the use cases including, safety, comfort, network congestion, demand and supply applications, navigations, security, and vehicle platoons. This work provides limited discussion on ML-based attack detection/prevention techniques. Hossain et. al \cite{nRef9SURVEY} present a comprehensive survey of ML approaches in vehicular applications. However, the scope of this survey is limited to cognitive radio-based (CR-based) vehicular networks. This work considers the amalgamation of ML in CR-VANET as a major research domain in the near future. This survey presents an overview of ML, VANET, and CR. It briefly describes the application and use of ML methods in various aspects of CR-VANETs. It includes spectrum sharing, spectrum mobility management, security issues, road safety, traffic congestion, resource allocation, spectrum-aware routing, and infotainment. This work addresses security threats only very briefly, focusing on the attacks in the CR environment. Veres et. al \cite{DLits} explore a wide range of ML-based architectures and their use in transportation networks. In \cite{DLits}, several network dynamics are investigated that include, destination prediction, demand prediction, traffic flow prediction, travel time estimation, transportation mode prediction, traffic signal control, navigation, demand serving, combinatorial optimization, and so on. This work does not address the problem of transportation security or the use of ML in this domain. \par 
Haydari et. al \cite{NS_DRLTrans} survey the use of reinforcement learning (RL) and DRL in optimal traffic signal control, autonomous driving, energy management, road control, and other intelligent transportation system (ITS) applications. This work does not cover the area of security. A security, privacy, and trust-focused review paper is presented in \cite{NS_SPT_MIoT}. This paper surveys a broad domain of mobile-IoT applications in which vehicular networks are one of such applications. In addition, ML-based solutions are not widely discussed in this work. Similar to this, another work on IoT security is presented in \cite{NS_IoTSec} with a little focus on ML-based solutions. There are some works in the literature which consider ML and deep learning (DL) in the survey on IoT security \cite{NS_MLIoTSec1,NS_MLIoTSec2}. In \cite{NS_MLIoTSec1}, authors briefly discuss security threat types and threats surface in IoT applications. A layer-based classification of literature on IoT security is presented with a very limited discussion on transportation/vehicular applications. Similar to \cite{NS_MLIoTSec1}, \cite{NS_MLIoTSec2} also present layer-based attacks in IoT. The scope of ML and DL solutions in this work includes malware analysis, authentication, intrusion detection, and attack detection. This work also covers a wide range of applications of IoT and a limited discussion on transportation applications. A vehicular communication-specific work for security, privacy, and trust management is presented in \cite{NS_RecentV2X}, but ML-based solutions are not the focus of this work. \par 
Recently, a survey on federated learning in vehicular-IoT is presented in \cite{FLPS2}. This work briefly introduces federated learning, its properties, and provides a comparison from other learning types. This work also reports the literature on using FL in other wireless IoT applications which can support vehicular usage. The existing research related to FL in vehicular IoT is classified into three different layers including, perception, networking, and application layes. This work studies a wide range of applications with a limited focus on the security and privacy of vehicular IoT networks. However, this work considers the use of FL as a promising research direction for future vehicular applications in terms of security, privacy, and incentive. Kuutti et. al \cite{AD1} recently present a deep learning (DL)-focused survey on the use of intelligent mechanisms in vehicle control systems. This work carries out a thorough work on the use of DL for different types of control systems in vehicular networks. It considers the use of DL in vehicular networks as promising that could achieve excellent performance. The works cited therein handle safety issues along with control problems. However, this survey does not provide any discussion on security problems.

\subsection{Our Contribution}
Different from the above-mentioned surveys, we present an in-depth study of the state-of-the-art on the use of ML techniques for security in vehicular networks. A wide range of security problems that include attacks, privacy, trust, intrusion detection, and driver identification/fingerprinting are discussed in this paper. Our work focuses on vehicular network security that comprehensively summarizes the works \textit{specifically} from the perspective of ML-based solutions. We first present the basics of vehicular networks and their variants with the adoption of various technologies such as 5G, SDN, IoT, edge computing, and cloud computing that enable intelligent transportation applications. We present security challenges and requirements for vehicular networks. We propose a taxonomy of attacks at different levels of vehicular networks. The significance of different ML techniques such as DL, RL, transfer learning (TL), and federated learning (FL) are elaborated in this survey. We summarize and contrast each work and present a holistic view in the form of tables for different security issues. We present the limitations and challenges, in using ML techniques for vehicular security. We summarize the lessons learned to provide useful insights.

\subsection{Structure of our Survey Work}

\begin{figure}[htbp]
	\begin{center}
		\begin{forest}
			for tree={
				grow'=0,
				s sep=0.5mm,
				child anchor=west,
				parent anchor=south,
				anchor=west,
				calign=first,
				edge path={
					\noexpand\path [draw, \forestoption{edge}]
					(!u.south west) +(6.5pt,0) |- node[fill,inner sep=0.5pt] {} (.child anchor)\forestoption{edge label};
				},
				before typesetting nodes={
					if n=1
					{insert before={[,phantom]}}
					{}
				},
				fit=band,
				before computing xy={l=12pt},
			}
			[\textbf{STRUCTURE OF THIS SURVEY}
			[\readded{\textbf{\textcolor{teal}{Section I}} Introduction} 
			[\readded{\textbf{\textcolor{gray}{Section IA}} Related Surveys}]
			[\readded{\textbf{\textcolor{gray}{Section IB}} Our Contribution}]
			[\readded{\textbf{\textcolor{gray}{Section IC}} Structure of our Survey}]
			]
			[\readded{\textbf{\textcolor{teal}{Section II}} Vehicular Networks} 
			[\readded{\textbf{\textcolor{gray}{Section IIA}} Vehicular Network Architecture} ]
			[\readded{\textbf{\textcolor{gray}{Section IIB}} Vehicular Networks and its Variants} ]
			[\readded{\textbf{\textcolor{gray}{Section IIC}} Types of Communication} ]
			[\readded{\textbf{\textcolor{gray}{Section IID}} Security Attacks and Requirements} ]
			]
			[\readded{\textbf{\textcolor{teal}{Section III}} A Brief Overview of Machine Learning} 
			[\readded{\textbf{\textcolor{gray}{Section IIIA}} Supervised Learning} ]
			[\readded{\textbf{\textcolor{gray}{Section IIIB}} Unsupervised Learning} ]
			[\readded{\textbf{\textcolor{gray}{Section IIIC}} Reinforcement Learning} ]
			[\readded{\textbf{\textcolor{gray}{Section IIID}} Deep Learning} ]
			[\readded{\textbf{\textcolor{gray}{Section IIIE}} Federated Learning} ]
			[\readded{\textbf{\textcolor{gray}{Section IIIF}} Transfer Learning} ]
			]
			[\readded{\textcolor{teal}{Section IV} ML-based Security Solutions}
			[\readded{\textbf{\textcolor{gray}{Section IVA}} Driver Identification/Fingerprinting}]
			[\readded{\textbf{\textcolor{gray}{Section IVB}} Attack Detection}]
			[\readded{\textbf{\textcolor{gray}{Section IVC}} Misbehaviour or Intrusion Detection}]
			[\readded{\textbf{\textcolor{gray}{Section IVD}} Trust Computation}]
			[\readded{\textbf{\textcolor{gray}{Section IVE}} Privacy Protection}]
			]
			[\readded{\textcolor{teal}{Section V} Limitations, Challenges and Open Issues}
			[\readded{\textbf{\textcolor{gray}{Section VA}} Adversarial Machine Learning} ]
			[\readded{\textbf{\textcolor{gray}{Section VB}} Latency Limitation in ML-based Solutions} ]
			[\readded{\textbf{\textcolor{gray}{Section VC}} Energy Constraint in ML-based Solutions} ]
			[\readded{\textbf{\textcolor{gray}{Section VD}} Computation-cost in ML-based Solutions} ]
			]
			[\readded{\textcolor{teal}{Section VI} Observations and Lessons Learned}]
			[\readded{\textcolor{teal}{Section VII} Conclusion}
			]
			]
		\end{forest}
		\caption{Structure of this Survey}
		\label{fig:structure}
	\end{center}
\end{figure}
The rest of this survey is organized as follows (as shown in Fig. \ref{fig:structure}). First, it describes the basics of vehicular networks in Section \ref{Sec:VN} that includes network architecture, variants of vehicular networks, types of communication used for transfer of data between different entities, and taxonomy of security attacks and requirements. The detailed classification of widely-used ML approaches in vehicular network security is explained in Section \ref{Sec:ML}. Section \ref{Sec:LearningSolutions} reviews the literature employing ML strategies for providing different security and privacy solutions. It briefly explains and compares different solutions and provides a tabular form of solutions. Section \ref{Sec:MLlimitations} highlights the limitations and challenges in using ML-based solutions for vehicular network security. Section \ref{Sec:Observations} presents observations and lessons learned from the works presented in this survey. Finally, we conclude our work in Section \ref{Sec:Conclusion}.

\section{Vehicular Networks}
\label{Sec:VN}
In this section, we describe the basic architecture (Section \ref{sec:basics}) of vehicular networks and its usage with different related fields (Section \ref{sec:ntwtrend}). We discuss different communication methods (Section \ref{sec:communication}) used between different entities to provide connectivity in vehicular networks. We then present different security attacks and requirements (Section \ref{sec:security}) in vehicular networks. 

\subsection{Vehicular Network Architecture}
\label{sec:basics}
\begin{figure}[htbp]
	\centerline{\includegraphics[width=3.3in, height=3.1in]{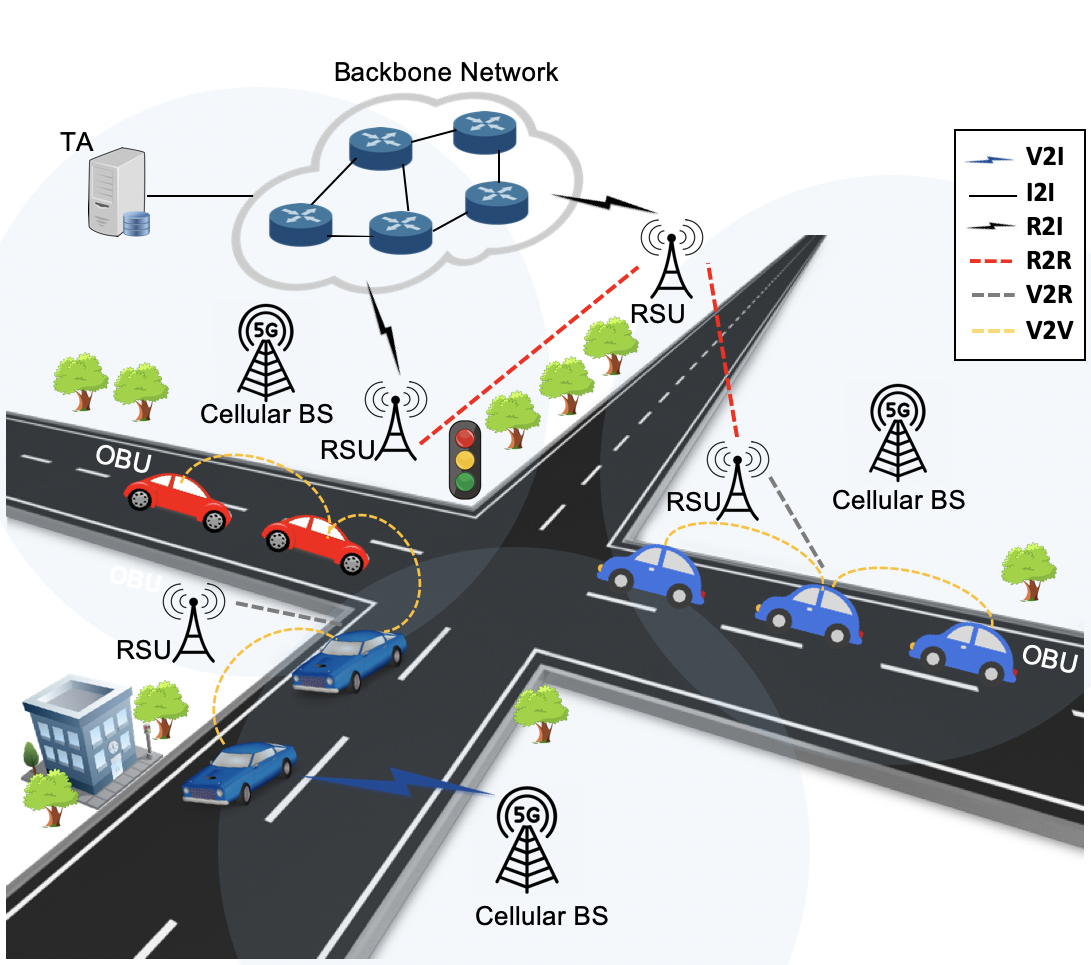}}
	\caption{An illustrative architecture of vehicular networks.}
	\label{fig:archi}
\end{figure}

The traditional architecture of vehicular networks helps to assist autonomous and non-autonomous vehicles. It is comprised of different components that include OBU, RSU, cellular base-station (BS), backbone network, and TA, as shown in Fig. \ref{fig:archi}. The OBUs are installed on the vehicles with necessary components for sensing different vehicular parameters such as speed, velocity, location coordinates, and proximity with different objects/vehicles. The RSU and/or cellular BS serve as an interface for vehicles with the backbone network. To transfer data from a vehicle to RSU, different wireless protocols can be used. Among them, the most common protocols used for short-range communication in vehicular applications are DSRC (Dedicated Short Range Communications) \cite{DSRC} and WAVE (Wireless Access in vehicular Environments) IEEE 802.11p \cite{IEEE80211}. A cellular BS is used when a vehicle is far from the range of DSRC/WAVE. The RSU connects with different components of the backbone network using wired or wireless connectivity. This results in different kinds of communications within a network which are explained in Section \ref{sec:communication}. The key function of TA is to use basic authorization techniques for the vehicles willing to register within a network.

\subsection{Evolution of Vehicular Networks and its Variants}
\label{sec:ntwtrend}
\begin{figure}[h]
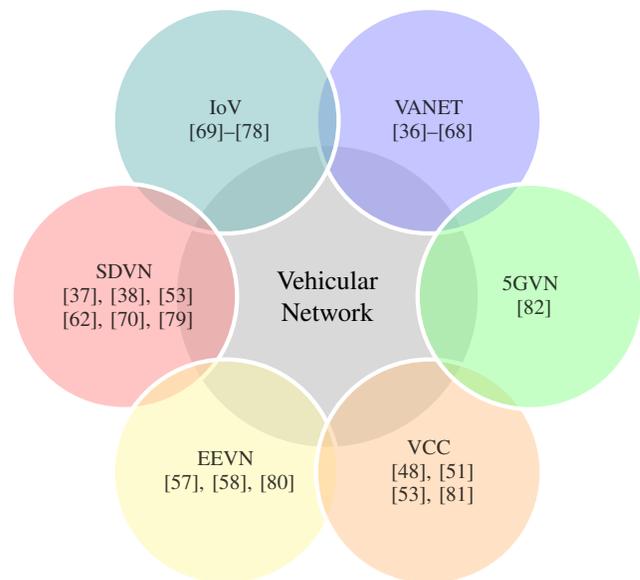

	\centering
			\smartdiagramset{
				bubble node size =3cm, 
				bubble center node font = \normalsize,
				bubble node font = \footnotesize, 
				distance center/other bubbles = 0.7cm, 
				set color list={blue!45, teal!55, red!45, yellow!45 ,orange!45,
					green!45},
				uniform connection color=true
			}%
			\smartdiagram[bubble diagram]{Vehicular \\Network,
				VANET \\\cite{ref22,ref25,ref46,ref26,ref32,ref27,ref30,ref33,ref38,ref50,ref31,ref34,ref39,ref43,nRef4,ref45,ref47,ref48,nRef1,ref51,ref54,ref55,ref57,ref58,ref59,ref67,ref68,ref69,ref79, ref81,nRef5,nRef7,SupMDS}, IoV \\ \cite{ref40,ref53,ref70,ref80,TLIDS,USAttack,PPFL,MLDCMDS,RLDOS,FLSecureData}, SDVN \\ \cite{ref25,ref46,ref48} \\ \cite{ref53,ref68,nRef6}, EEVN \\ \cite{ref55,ref57,FLPP2}, VCC \\ \cite{ref39,ref45} \\\cite{ref48,nRef2}, 5GVN \\ \cite{nRef8}}
	\caption{Variants of vehicular networks and related work.}
	\label{fig:trends}
\end{figure}

With the technological advances such as IoT and 5G, vehicular networks have gone through an evolutionary process from VANET to IoV\cite{IoV}. IoV provides intelligence by integrating an environmental understanding of surrounding things such as human (driver) actions and activities. This results in a new level of communication known as Vehicle-to-Person (V2P) (explained in Section \ref{sec:communication}). Further, the emergence of modern technologies such as 5G, SDN, edge computing, and cloud computing has enabled different applications leading to the creation of new variants of vehicular networks, as shown in Fig. \ref{fig:trends}. \par

5G technology that could provide high speed and low latency has become a key driver for VANETs in enhancing transportation efficiency \cite{5GVANET}. It is very crucial in vehicular applications such as fully-autonomous driverless cars to transmit data with a delay of less than 1ms to make correct driving decisions. The standard VANET wireless protocol IEEE 802.11p has the problem of intermittent connectivity and poor capacity for the rigorous autonomous vehicles \cite{5GSDVN}. This has led to proposals to integrate 5G with vehicular networks which are known as 5GVN. Apart from low delay, 5G-integrated vehicular networks can provide efficient solutions for congestion control, fair resource sharing, reliability, high-throughput, high-connectivity and support diverse safety applications \cite{5GVANET}. \par 
SDN is a key enabler for creating an effective topology in vehicular networks \cite{SDVN}. SDN simplifies hardware-software management through a flexible networking architecture to effectively handle the most dynamic nature of the vehicles. In the literature, there have been several studies on integrating SDN at different levels of the network and the design of a hierarchical and centralized Software-Defined Vehicular Network (SDVN) to support different functionalities \cite{5GSDVN} \cite{SDVN}. The use of SDN helps to enhance Quality of Service (QoS), routing reliability, and security services of the nodes in vehicular networks. Most importantly, the use of SDN at the edge of vehicular networks is effective for scheduling valuable data packets and achieving improved QoS. This edge-enabled vehicular network (EEVN) solutions are also ideal for low-delay applications. However, the use of SDN at the edge is not ideal for all applications. In the case of many vehicles connected to a single RSU, it can result in performance degradation of SDVN due to frequent handovers \cite{SDVNSch} \cite{handover}. In the literature, there exist several proposals that focus more on the use of a centralized controller in SDVN architectures to achieve better security, improved scalability, and traffic management \cite{SDVN}. Nonetheless, controller placement has always been a major issue \cite{ControllerPlacement} which demands more research in the direction of vehicular applications. \par
Some recent works exploit the idea of minimizing onboard storage and computation in vehicular networks by using cloud-based services \cite{cloudVANET}\cite{cloudVANET2}. Cloud computing is a potential technology to provide flexible solutions by providing access to virtual services to road users. This results in a new variant of vehicular networks, known as Vehicular Cloud Computing (VCC). The first cloud-based VANET architecture was proposed in \cite{cloudVANET}. This work discusses several issues including security and privacy in vehicular networks using VCC design. \par
While the adoption of the above key technologies brings benefits, they also pose challenges. The connectivity of vehicles to everything using advanced technologies creates serious safety concerns. In vehicular networks, these advanced variants can be used separately or with each other to provide enhancements on scalability, power efficiency, spectrum efficiency, and flexibility. An important problem with all these technologies is to ensure security and privacy within a network. Our survey mainly focuses on ML-based security solutions for vehicular networks. Fig. \ref{fig:trends} shows the works proposing ML-based security solutions in different variants of vehicular networks.

\subsection{Types of Communication}
\label{sec:communication}

\begin{table}[h]
  \centering
  \caption{ML-based Literature to Detect or Prevent Adversaries at Different Levels of Communications}
    \begin{tabular}{lL{2.7cm}rrcP{1cm}}
    \toprule
    \multicolumn{1}{c}{\multirow{2}[2]{*}{\textbf{Citation}}} & \multicolumn{1}{c}{\multirow{2}[2]{*}{\textbf{Adversary Model}}} & \multicolumn{3}{c}{\textbf{Communication}} & \multicolumn{1}{c}{\multirow{2}[2]{*}{\textbf{In-Vehicle}}} \\
          &       & \multicolumn{1}{c}{\textbf{V2V}} & \multicolumn{1}{c}{\textbf{V2I}} & \textbf{V2R} &  \\
    \midrule
    \cite{ref22} & Platoon Attack & \multicolumn{1}{c}{\cmark} &       &       & \multicolumn{1}{c}{\cmark} \\
    \cite{ref25} & DDoS Attack  &       & \multicolumn{1}{c}{\cmark} &       &  \\
    \cite{ref46} & DDoS Attack  & \multicolumn{1}{c}{\cmark} & \multicolumn{1}{c}{\cmark} &       &  \\
    \cite{USAttack} & DDoS Attack  &  &  &       &  \multicolumn{1}{c}{\cmark} \\
    \cite{RLDOS} & DDoS Attack & \multicolumn{1}{c}{\cmark} & \multicolumn{1}{c}{\cmark} &       &  \multicolumn{1}{c}{\cmark} \\
    \cite{ref26} & GreyHole \& BlackHole &       &       & \cmark   &  \\
    \cite{ref32} & Black hole & \multicolumn{1}{c}{\cmark} &       &       &  \\
    \cite{ref27} & Sybil Attack &       & \multicolumn{1}{c}{\cmark} &       & \multicolumn{1}{c}{\cmark} \\
    \cite{ref30} & Sybil Attack & \multicolumn{1}{c}{\cmark} & \multicolumn{1}{c}{\cmark} &       &  \\
    \cite{ref33} & Sybil Attack & \multicolumn{1}{c}{\cmark} & \multicolumn{1}{c}{\cmark} & \cmark   &  \\
    \cite{ref38} & Jamming Attack & \multicolumn{1}{c}{\cmark} &       &       &  \\
    \cite{ref40} & Jamming Attack & \multicolumn{1}{c}{\cmark} &       &       &  \\
    \cite{ref50} & jamming Attack & \multicolumn{1}{c}{\cmark} &       &       &  \\
    \cite{ref24} & Data Manipulation & \multicolumn{1}{c}{\cmark} &       &       & \multicolumn{1}{c}{\cmark} \\
    \cite{ref53} & Crossfire Attack &       & \multicolumn{1}{c}{\cmark} &       &  \\
    \cite{ref54} & Spoofing Attack &       &       & \cmark   &  \\
    \cite{ref55} & Spoofing Attack &       &       & \cmark   &  \\
    \cite{ref57} & Spoofing Attack &       &       & \cmark   &  \\
    \cite{RLCPAttack} & Cyber Physical Attack &       &       &   &  \cmark \\
    \cite{DLCPAttack} & Cyber Physical  Attack &       &       &   &  \cmark  \\
    \cite{ref31} & MDS   & \multicolumn{1}{c}{\cmark} &       &       &  \\
    \cite{ref34} & MDS   & \multicolumn{1}{c}{\cmark} & \multicolumn{1}{c}{\cmark} & \cmark   & \multicolumn{1}{c}{\cmark} \\
    \cite{ref43} & MDS   & \multicolumn{1}{c}{\cmark} & \multicolumn{1}{c}{\cmark} &       &  \\
    \cite{nRef4} & MDS   & \multicolumn{1}{c}{\cmark} &  	 &       &  \\
    \cite{nRef8} & MDS   & \multicolumn{1}{c}{\cmark} & \multicolumn{1}{c}{\cmark} & \multicolumn{1}{c}{\cmark}     &  \\
    \cite{SupMDS} & MDS   & \multicolumn{1}{c}{\cmark} & \multicolumn{1}{c}{\cmark} &   &  \\
    \cite{MLDCMDS} & MDS   & \multicolumn{1}{c}{\cmark} &  &   &  \\
    \cite{ref51} & MDS   & \multicolumn{1}{c}{\cmark} &       & \cmark   &  \\
    \cite{ref39} & FDI & \multicolumn{1}{c}{\cmark} & \multicolumn{1}{c}{\cmark} &       &  \\
    \cite{ref45} & IDS   & \multicolumn{1}{c}{\cmark} &       & \cmark   &  \\
    \cite{ref47} & IDS   & \multicolumn{1}{c}{\cmark} &       &       &  \\
    \cite{ref58} & IDS   & \multicolumn{1}{c}{\cmark} &       & \cmark   &  \\
    \cite{ref59} & IDS   & \multicolumn{1}{c}{\cmark} &       & \cmark   &  \\
    \cite{ref48} & IDS   & \multicolumn{1}{c}{\cmark} & \multicolumn{1}{c}{\cmark} &       &  \\
    \cite{nRef1} & IDS   & \multicolumn{1}{c}{\cmark} &  	&       &  \\
    \cite{nRef2} & IDS   & 		& \multicolumn{1}{c}{\cmark} &       &  \\
    \cite{TLIDS} & IDS   & 		& \multicolumn{1}{c}{\cmark} &       &  \\
    \cite{CANTLIDS} & IDS   & 		&  &       &  \multicolumn{1}{c}{\cmark}\\
    \cite{ref61} & Trust Computation & \multicolumn{1}{c}{\cmark} &       &       &  \\
    \cite{ref66} & Trust Computation & \multicolumn{1}{c}{\cmark} &       & \cmark   &  \\
    \cite{ref67} & Trust Computation & \multicolumn{1}{c}{\cmark} &       & \cmark   &  \\
    \cite{ref68} & Trust Computation & \multicolumn{1}{c}{\cmark} & \multicolumn{1}{c}{\cmark} &       &  \\
    \cite{ref69} & Trust Computation & \multicolumn{1}{c}{\cmark} &       &       &  \\
    \cite{ref70} & Trust Computation & \multicolumn{1}{c}{\cmark} &       &       &  \\
    \cite{ref81} & Trust Computation &       &       & \cmark   & \multicolumn{1}{c}{\cmark} \\
    \cite{nRef5} & Trust Computation &  &  \multicolumn{1}{c}{\cmark}     &       &  \\
    \cite{nRef6} & Trust Computation & \multicolumn{1}{c}{\cmark} &       &       &  \\
    \cite{nRef7} & Trust Computation & \multicolumn{1}{c}{\cmark} &       &       &  \\
    \cite{ref79} & Privacy Protection &       &       & \cmark   &  \\
    \cite{ref80} & Privacy Protection &       &       & \cmark   &  \\
    \cite{PPFL} & Privacy Protection   & \multicolumn{1}{c}{\cmark} & \multicolumn{1}{c}{\cmark} &   &  \\
    \cite{FLSecureData} & Privacy Protection &       &       & \cmark   &  \\
    \cite{FLPP2} & Privacy Protection &       &  \cmark      & \cmark   &  \\
    \bottomrule
    \end{tabular}%
  \label{tab:CommMode}%
\end{table}%
Data transfer in Vehicle-to-Everything (V2X) involves different types of communications at different levels. The first type is low-level communication which involves the transfer of data between the end nodes such as V2V, V2P, and Vehicle-to-Sensor (V2S). In V2V communication, messages related to traffic congestion and accidents are transmitted between different regions without involving the backbone network. On the other side, V2P and V2S modes of communication help to integrate the conditions of persons and sensors while making decisions such as driving. This level of communication mostly uses Bluetooth, DSRC, and IEEE 802.11p protocols.\par  
The next type of communication takes place between a vehicle and an intermediate node (RSU), or between two intermediate nodes, to carry out operations such as broadcasting, routing, vehicle authorization, and access services from the server. This includes V2R and RSU-2-RSU communication (R2R). This level of communication uses short-range communication if RSU is closer and LTE/5G in case of long-distance. The third type is high-level communication which is used to transfer packets from an RSU to backbone infrastructure (R2I) and I2I. This type of communication is required when a vehicle requests a service from the backbone network such as LBS, infotainment service, entertainment service, and also for generating certificates from TA and centralized routing decisions. This includes long-range wireless communication or wired communication.\par
Another important mode of communication is V2I in which a vehicle is smart enough to communicate to the infrastructure directly using cellular services (LTE/5G) for faster decisions and packet transfers without an intermediate node. However, the improved connectivity and increased number of communication channels have led to several breakthroughs and security problems. Such threats at different levels of communications have been studied and ML-based solutions have been developed in the literature. We list these works in Table \ref{tab:CommMode}. We note that most of the ML-based works in the literature focus on security in V2V and V2R communication. Recently, the advent of autonomous vehicles has driven research on security of in-vehicle network. Furthermore, advances in 5G and SDN facilitate V2I communication and there is ongoing research in securing this communication as well.

\subsection{Security Attacks and Requirements}
\label{sec:security}
\begin{figure}[htbp]
	\centerline{\includegraphics[width=3.5in, height=2.7in]{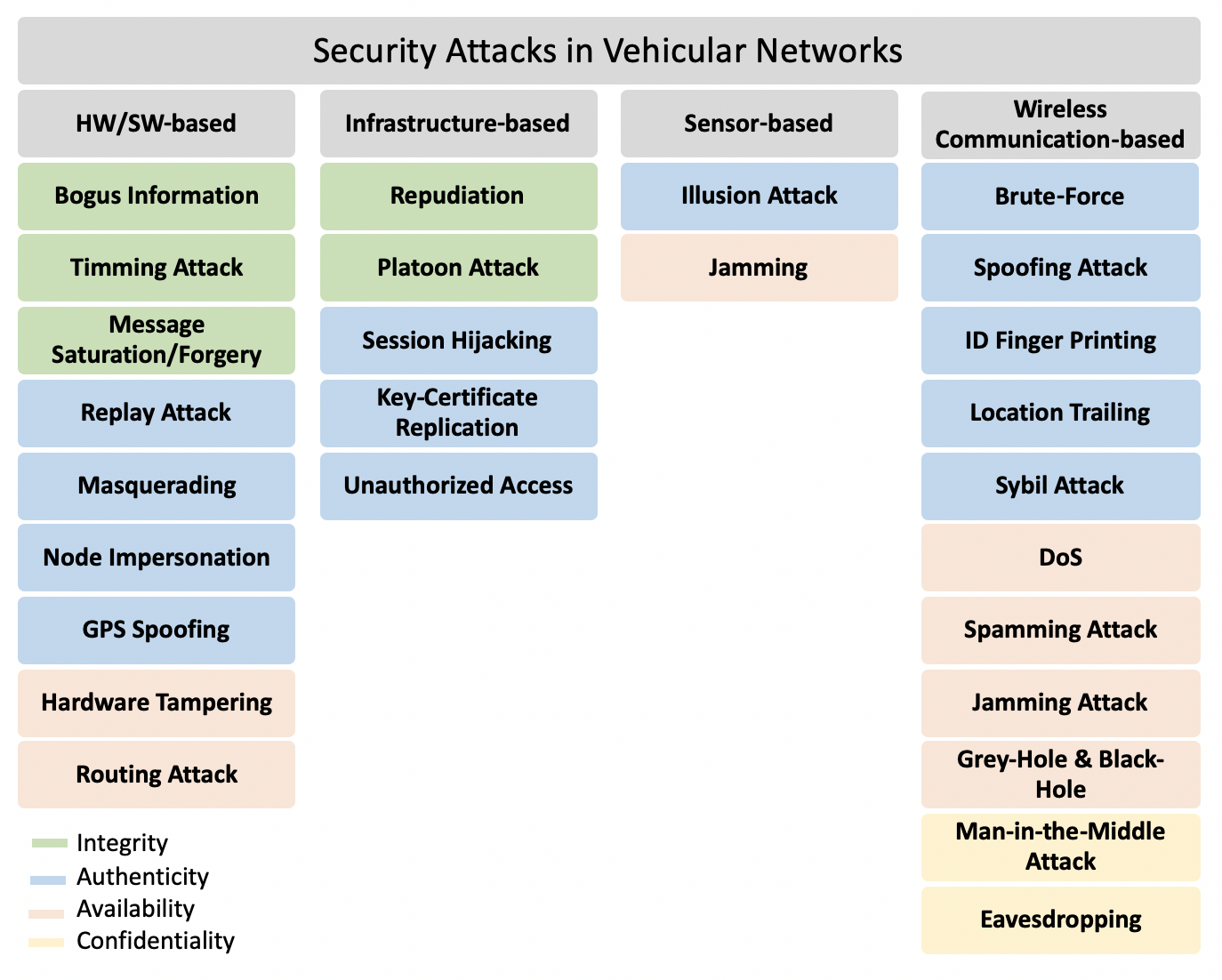}}
	\caption{Taxonomy of attacks in vehicular network.}
	\label{fig:attacks}
\end{figure}

This section addresses the requirements and threats related to security in vehicular networks. We present a taxonomy of security attacks at different parts of the vehicular network as shown in Fig. \ref{fig:attacks}. Accordingly, we classify the attacks into four (4) different classes, i.e. hardware or software (HW/SW)-based, infrastructure-based, sensor-based, and wireless communication-based. In Fig. \ref{fig:attacks}, we depict the possible security attacks in the above four classes along with the security requirements highlighted using different colors. 

\subsubsection{Attacks}
We briefly describe the attacks that are categorized under four classes in this section. 

\begin{description}

	\item \textbf{Hardware/Software-based Attacks} Several attacks can take place over the hardware components and software systems of the VANET network to compromise different security requirements. Some of the common attacks are listed below.

	\begin{description}

		\item \textit{Bogus Information}:  In this type of attack, the attacker sends a piece of bogus or false information to misguide the functioning of hardware or software systems in the vehicles. 

		\item \textit{Timming Attack}: This is a side-channel attack. It tries to compromise the cryptographic algorithm of a system by analysing its timing information which is required to execute the attack. 

		\item \textit{Message Forgery}: This type of attack is launched to deceive the recipient about the real sender.

		\item \textit{Replay Attack}: This attack is also known as playback attack, where transmission of valid or true data is repeated or delayed maliciously to misguide the functioning of the system. 

		\item \textit{Masquerading Attack}: It is a form of attack in which the attacker uses fake identification to gain unauthorized access to the vehicle system.  

		\item \textit{Node Impersonation:} In this type of attack, the attacker steals the identity of an authorized user to gain access to the system.

		\item \textit{GPS Spoofing}: This attack tries to fool a global positioning system (GPS) by generating fake signals around the vehicle in which the GPS sensor captures the fake signals and records fake coordinates to misguide the functioning of the vehicular networks. 

		\item \textit{Tampering Hardware}: It is a type of attack in which the hardware of a vehicle is deceived by providing fake information or creating a fake environment around the vehicle. 

		\item \textit{Routing Attack}: This attack results in improper functioning of the routing process. The routing attacks are further classified into different attacks corresponding to malfunctioning of the routing process at different levels in a network. In terms of hardware, miscommunication of the node's presence and location may affect the routing tables in a network.   \\

	\end{description}

	\item \textbf{Infrastructure-based Attacks} This section describes the attacks which infect the system at the infrastructure level. 

	\begin{description}

		\item \textit{Repudiation Attack}: This attack takes place at the application layer where a system fails to control the log of actions and tracking of nodes due to malicious manipulations. It is also known as the act of refusing the actions in a system. 

		\item \textit{Platoon Attack}: Platoon is a concept of grouping vehicles that travel in the same lane with close proximity and similar speed regulations. Any type of action by an attacker to destabilize the functioning of the platoon is known as a platoon attack.   

		\item \textit{Session Hijacking}: Session Hijacking is a type of attack in which an attacker tries to hijack and get access to the session of data transfer established between the vehicle and destination node.

		\item \textit{Key-Certificate Replication}: In this attack, the attacker uses duplicate keys and certificates of legitimate users to fool TA and gains access to the network.  

		\item \textit{Unauthorized Access}: This is an attack over the authentication systems of the vehicular networks. In this attack, the attacker tries to get access to the vehicle node system, server, controller, or any other component of the network by compromising its authentication parameters such as decrypting login identification (ID) and password for a system account.  \\

	\end{description}

	\item \textbf{Sensor-based Attacks} These attacks concern about compromising the authenticity and availability of sensors over the vehicles. There are two such attacks in vehicular networks which are described below.
			\begin{description}
				\item \textit{Illusion Attack}: In this attack, an adversary vehicle deceives its own sensors to produce wrong readings and transmits to the network. This creates a fake illusion of the scenario over the road to misguide other vehicles and generates false warning messages. 
				\item \textit{Jamming Attack}: In this attack, the aim of the jammer is to block or interfere transmission of data from a sensor by sending false alerts or creating a spoofed environment around the sensor. \\
			\end{description}
	\item \textbf{Wireless Communication-based Attacks} Wireless communication channel is one of the most vulnerable targets in a network. A large variety of attacks can take place over the wireless channel to compromise different security requirements which are discussed below.

	\begin{description}

		\item \textit{Brute-Force Attack}: It is an attack over the authentication system of a wireless protocol used for transferring data in V2X communication. The adversary runs a brute-force algorithm to create different combinations of passwords or pass-phrases to break access into the medium. 

		\item \textit{Spoofing Attack}: In vehicular applications, a different type of spoofing attack can take place wherein the attacker pretends as a legitimate user of the network to gain access over the personal information. A spoofing attack is not only limited to the spoofing of identification but also the location, domain-name-server (DNS) information, internet protocol (IP) address, and so on. 

		\item \textit{ID Fingerprinting}: In a vehicular network, it is of utmost importance to have correct identification and true profiling of the driver to prevent hacking and theft of the car. In ID fingerprinting, the attacker aims to obtain the driver profile and uses it to launch an attack over the system. 

		\item \textit{Location Trailing}: This attack violates privacy by illegally getting access to the channel that is transmitting the vehicle's personal information. Here, an adversary can track the complete path of the target vehicle and follows its location wherever it goes. 

		\item \textit{Sybil Attack}: Sybil attack is one of the most common and easy-to-implement attacks. Sybil attack in a vehicular network creates virtual nodes to launch an attack and detection of such virtual nodes is not easy. 

		\item \textit{Denial of Service (DoS)}: This is one of the difficult-to-handle and very frequently implemented attacks. In this attack, the attacker launches a bulk of spoofed requests over the server or any other node to make it fully occupied with unnecessary requests and blocks the access to legitimate users.

		\item \textit{Spamming Attack}: In this attack, the attacker sends the bulk of spam messages to consume network bandwidth and increases delay for the transmission of data.  

		\item \textit{Jamming Attack}: In this attack, the aim of the jammer is to block or interfere the authorized transmission by occupying the channel or by sending false alerts. 

		\item \textit{Grey-Hole and Black-Hole Attack}: The black-hole and grey-hole attacks are types of wireless routing attacks. In these attacks, a node tries to stop onward forwarding of messages/packets towards the receiver. In a black-hole attack, there will be a complete blackout or drop of packets. However, in a grey-hole attack, a partial drop of packets will take place and partial packets are altered by an attacker to convey wrong information to the receiver.

		\item \textit{Man-in-the-Middle (MiTM) Attack}: In MiTM, an adversary hears the communication going-on between two nodes through intercepting the channel and pretends as one of them to reply with the wrong information. 

		\item \textit{Eavesdropping}: It is a sniffing attack, where the attacker snoops the information transmitted between two entities. In this attack, it does not alter or reply to any of the entities but only listens and gains access to personal information. 

	\end{description}

\end{description}

\subsubsection{Requirements}
It is important to satisfy the basic security requirements to overcome the above security threats and attacks in vehicular networks. The taxonomy of security requirements for the VANET can be found in \cite{ref42}\cite{ref82}. The important security requirements for vehicular networks are summarized below.
\begin{description}

	\item \textbf{Confidentiality} One of the very basic requirements is to guarantee the access of data to legitimate users only. Confidentiality uses encryption techniques based on secure keys and trusted certificates to ensure access only to the legal user of data. Most of the works for confidentiality is about key management. If the attacker gets access to keys he can break into the system and misuses confidential data. Therefore, guaranteeing confidentiality is very important in vehicular networks to secure the personal data of vehicles and drivers. 

	\item \textbf{Availability} It is an important security requirement to assure the availability of the functionality of the applications and processes of any network. There are many critical attacks on the availability of services or communication channels in VANET which are hard to handle \cite{ref83}. Different types of attacks can take place at different levels of the network, as shown in Fig. \ref{fig:attacks}. It is very crucial for the successful running of the network to take prior steps and ensure the availability of the system if an unwanted situation arises. It is one of the basic security requirements to keep the communication between vehicles and infrastructure functional in the event of attacks. We discuss different ML-based solutions to detect attacks and ensure availability in Section \ref{Sec:LearningSolutions}. 	

	\item \textbf{Integrity} While transmitting data from a vehicle to infrastructure or vehicle to vehicle, it is important to maintain the true form of data and information. Any alteration or change in data may result in undesirable sequences. Integrity in vehicular networks is to ensure the originality of data and protect it from any change, destruction, or alteration from an adversary. The use of Public Key Infrastructure (PKI) and cryptographic schemes are generally used in the literature to ensure integrity in vehicular networks \cite{Ref3}. 

	\item \textbf{Privacy} Privacy is a critically-important requirement for security in vehicular networks. It is a means of protecting the sensitive information of a vehicle from the attacker. In the context of vehicular networks, it is further classified into location privacy and user privacy. The location and identity of the driver/vehicle are very sensitive and must be kept hidden from an attacker. However, trusted authorities must know the location information to provide better services. There are many LBSs which provide services to the users based on their location. In such cases, protecting the privacy of location from attackers while ensuring its secure availability to the service-providing entity is quite challenging \cite{ref82}. 

	\item \textbf{Authentication} Authentication is the first step of security in a vehicular network. A vehicle always authenticates itself to the system before requesting any service from it. There are two types of authentications, message authentication, and node authentication. When a vehicle becomes a part of any new network it authenticates its node information (i.e. identification, address and related information) with the network. This helps a system to differentiate between legitimate and malicious nodes. However, when a message is transmitted between two vehicles, it must be authenticated to ensure integrity. A failure to maintain the first step of security in a vehicular network may result in complete damage to the network. 

	\item \textbf{Trust} Trust is an important aspect of security to enhance the protection level of the system from attackers. Trust can be classified as a belief of one entity about another entity belonging to the same group of networks. In vehicular networks, trust computation is an additional step of security requirement used along with privacy protection, availability, and key management to ensure the highly-secure transmission of data \cite{ref84}. Trust computation schemes exist in the literature (in the context of entity-oriented and data-oriented) which make use of historic interaction of vehicles within a network to classify it as the trusted one \cite{ref85}. It is important in vehicular networks to ensure honesty among vehicles by performing trust computation as a basic security requirement. 	
\end{description}

\begin{figure}[htbp]
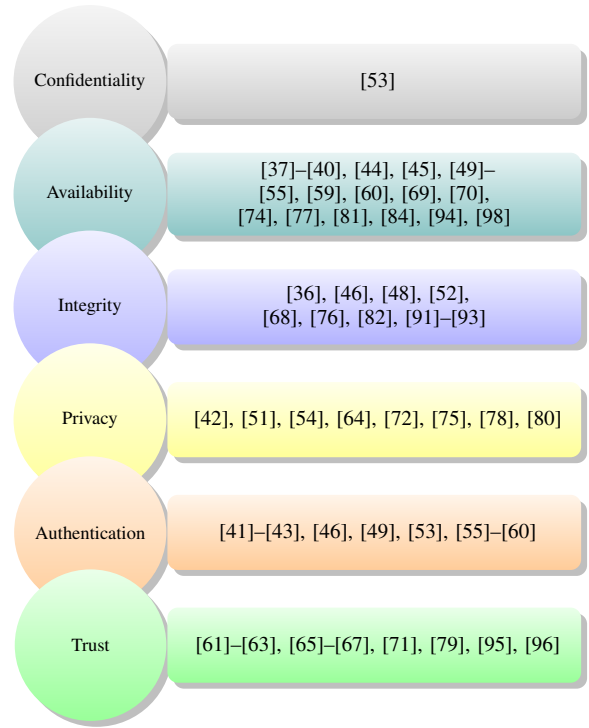

	\centering
	\smartdiagramset{descriptive items y sep=1.5cm , description title width=2cm, description title text width=1.75cm, description title font=\scriptsize, description font=\footnotesize, border color=none, set color list={gray!40, teal!45, blue!30, yellow!40 ,orange!40,
			green!40},
		uniform connection color=true
	}
	\smartdiagram[descriptive diagram]{
		{Confidentiality,{\cite{ref48}           }},
		{Availability     , {\cite{ref25,ref46,ref26,ref32,ref36,ref38,ref40,ref50,ref43,nRef4,ref45,ref47,ref48,nRef1,nRef2,5GSDVN,ref51,ref53,ref58,ref59,CANTLIDS,USAttack,RLDOS}}},
		{Integrity        , {\cite{ref22,ref24,ref31,ref39,nRef8,ref47,SupMDS,MLDCMDS,RLCPAttack,DLCPAttack}}},
		{Privacy          , {\cite{ref30,ref45,nRef1,ref79,ref80,PPFL,FLSecureData,FLPP2}}},
		{Authentication, {\cite{ref27,ref30,ref33,ref31,ref43,ref48,ref51,ref54,ref55,ref57,ref58,ref59}}},
		{Trust              , {\cite{ref61,ref66,ref67,ref68,ref69,ref70,ref81,nRef5,nRef6,nRef7}}}}
	\caption{Security requirements and ML-based literature. }
	\label{tab:SecREQ}
\end{figure}

This survey centers around the use of ML in achieving the above security requirements. As shown in Fig. \ref{tab:SecREQ}, a number of works have been carried out on the problems of availability, integrity, authentication, and trust computation. In the context of privacy, the use of ML is carried out in a very recent times. However, the use of ML for achieving confidentiality is not much studied in the literature. Most of the works in the literature are about the use of keys to secure data and achieve confidentiality within a system \cite{KeyManage1,KeyManage2,KeyManage3,KeyManage4}.

\section{A Brief Overview of Machine Learning in the Context of Vehicular Network Security}
\label{Sec:ML}
\begin{figure}[htbp]
	\begin{center}
		\includegraphics[width=2.8in,height=1.8in]{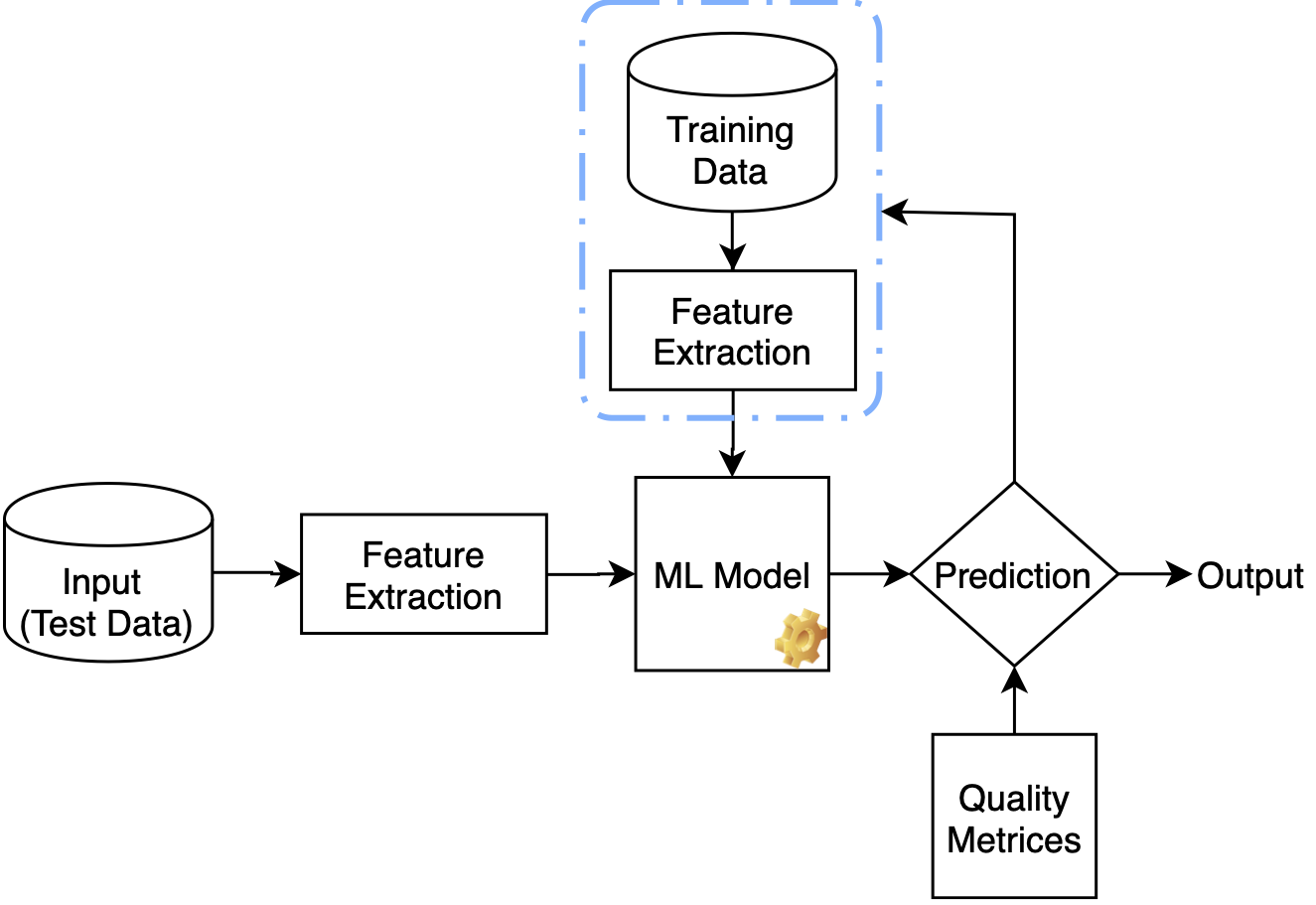}
		\caption{\readded{The conceptual block diagram of a typical ML model.}}
		\label{fig:MLModel}
	\end{center}
\end{figure}

ML is a branch of Artificial Intelligence (AI) proposed first in 1959 \cite{ML} as a self-learning technique for the game of checkers. Today, the use of ML is widely explored in almost all areas of networking \cite{MLSurvey}. ML is a computing-based strategy which determines the hidden insights of a dataset without being explicitly programmed. It improves the working performance from its learning-experience. A typical model for the traditional ML consists of three phases: 1) Training Phase, which takes raw data and pre-processes it to extract the features. The features are input into the ML model to learn patterns and classes of the data. 2) Testing Phase, where a new set of data is tested by the ML model for classification based on its learning-experience from the training phase. 3) Prediction Phase, also known as the evaluation phase where the working efficiency of an ML model is evaluated based on quality metrics (such as accuracy, false-positives, false-negatives, and so on). In the case of lower efficiency, the training phase updates its data and/or features for achieving better results. A typical ML model is shown in Fig. \ref{fig:MLModel}. \par
\begin{figure*}[t]
	\begin{center}
		\includegraphics[width=5.8in,height=4in]{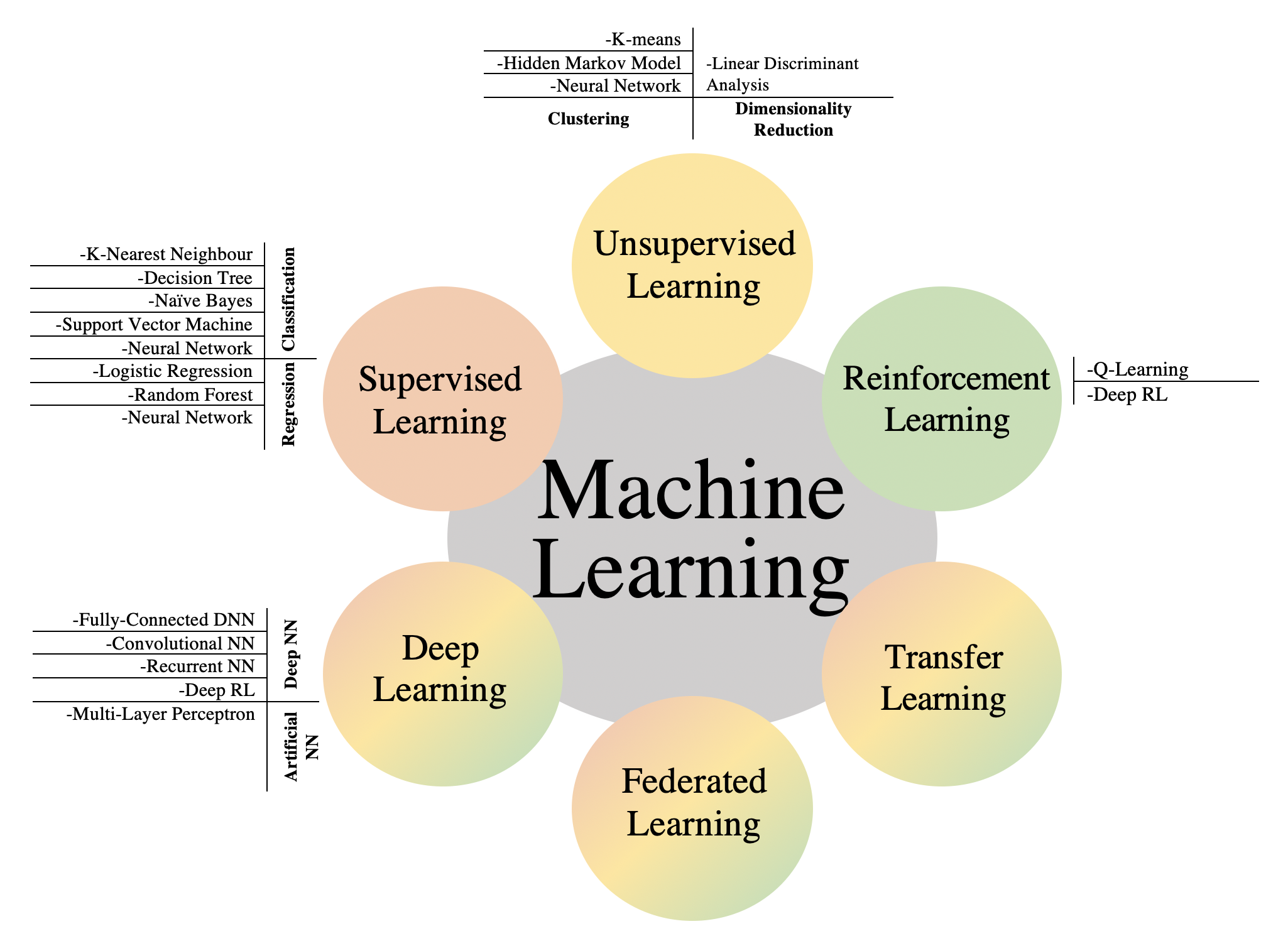}
		\caption{ML approaches in the context of vehicular network security.}
		\label{fig:LearningClassification}
	\end{center}
\end{figure*}
The ML techniques are further classified into three broad categories, namely, supervised, unsupervised learning and reinforcement learning (as shown in Fig. \ref{fig:LearningClassification}). The advances in the functionalities have evolved these classes into several other learning types such as DL, TL, and FL as shown in Fig. \ref{fig:LearningClassification}. These types work in parallel with the three main classes and have received significant attention due to their intelligence in performing a different kind of tasks. We explain below the widely-used ML approaches in vehicular network security.

\subsection{Supervised Learning}
In supervised learning, each entry of the training dataset consists of an input value and its corresponding label. The supervised algorithm learns the relation between the input sample and label of the training set and uses it to map the new instances of the testing data \cite{ref86}. Supervised learning can be applied for vehicular networks in different domains but our survey explores the use of supervised learning in securing vehicular networks.\par 
Supervised learning is further classified into classification and regression. The output of classification model is categorical or discrete. The commonly used classification models for the security in vehicular networks are K-Nearest Neighbour (KNN) \cite{NNbour} \cite{KNN}, Decision Tree \cite{DT}, Naive Bayes \cite{NB}, Support Vector Machine (SVM) \cite{SVM}, and Neural Network (NN) \cite{NNClassifier}. The output of regression model is a continuous value. The most-common regression models used to secure the vehicular networks are logistic regression \cite{LogisticR}, random forest \cite{RF}, and NN \cite{NNRegression}. In vehicular networks, the use of supervised learning applies in different applications such as driver fingerprinting, type of misbehaviour, attack detection, and trust computation (Section \ref{Sec:LearningSolutions}).

\subsection{Unsupervised Learning}
In contrast to supervised learning, unsupervised learning consists of input values only in their training dataset. There is no use of pre-assigned labels for the dataset in unsupervised learning. The idea of unsupervised learning is to find the hidden patterns of data from unlabeled information. As a result, similar structures of data are clustered into the same group. Unsupervised algorithms are efficient and faster in data processing. \par 
Unsupervised algorithms are classified into clustering and dimensionality reduction applications. In clustering, the input samples group together on the basis of different similarity attributes such as relative or absolute similarity. The grouping takes place by randomly selecting cluster centroids and similarity attributes toward all values are calculated from the center. The selection of centroid keeps on changing until the best match is found. The most common clustering mechanisms for the security in vehicular networks are k-means clustering \cite{Kmeans}, Hidden Markov Model (HMM) \cite{HMM}, and NN \cite{NNcluster}. In dimensionality reduction, the data is projected from a higher dimension to a lower dimension without losing useful information \cite{ref86}. The dimensionality reduction techniques are faster in optimization and less complex but may degrade the learning process. In terms of security in vehicular networks, the use of Linear Discriminant Analysis (LDA) as a dimensionality reduction mechanism is considered in the literature \cite{LDA}.

\subsection{Reinforcement Learning}
\begin{figure}[htbp]
	\begin{center}
		\includegraphics[width=2.5in,height=1.4in]{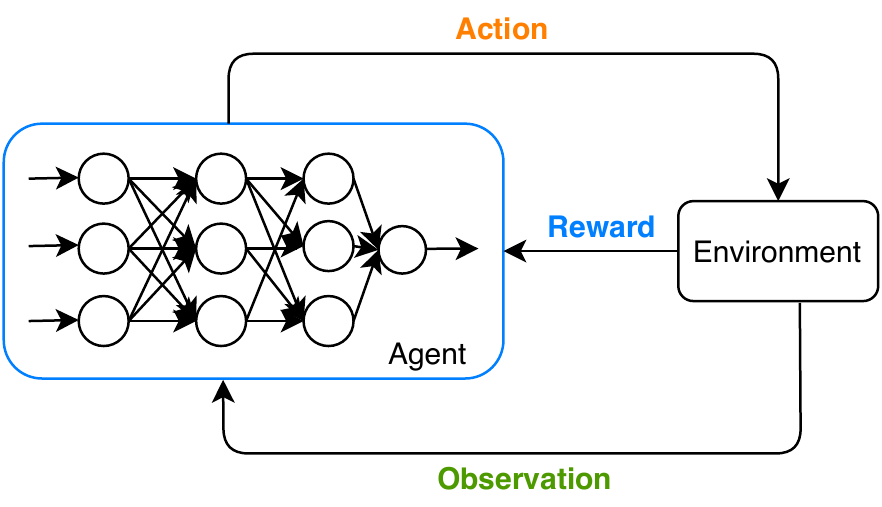}
		\caption{\readded{The conceptual framework of RL.}}
		\label{fig:RLModel}
	\end{center}
\end{figure}

Compared to supervised and unsupervised learning, RL uses different policies for the learning process. The conceptual framework of RL is shown in Fig. \ref{fig:RLModel}. The objective is to learn a policy which helps an agent to act optimally in the given environment \cite{RL}. An agent generates data to learn based on rewards received and tries to maximize the positive rewards by interacting with an environment using a trial-and-error method. The environment is a Markov decision process where the reward and state transition probability are defined by an observation and the selection of actions \cite{RL2}. The policy of RL is to find actions which maximize future rewards. \par 
One of the most popular and widely used RL methods is Q-learning. Q-learning uses the Bellman equation as a constraint to maximize the cumulative rewards \cite{RL}. It aims to maximize the expected sum of rewards by applying a policy for the selection of actions. In practice, Q-learning generates a lookup table to store a combination of actions and expected rewards. This requires larger memory and sometimes becomes inefficient when a continuous set of actions are used for the data. As a solution, the DRL network is proposed which involves the combination of DL and RL to handle larger datasets \cite{DQN}. The use of RL, DRL and its variants is widely exploited for security in vehicular networks, as explained in Section \ref{Sec:LearningSolutions}.

\subsection{Deep Learning}
DL is a subset of ML but works contrastingly to the traditional ML algorithms. It does not require feature engineering as in the traditional ML. In DL, it learns the pattern of data on its own through self-optimization of an algorithm \cite{DL3}. DL finds the patterns which are too hard for the traditional-ML to learn. Raw data, in its original form (without pre-processing), can directly be applied to the DL algorithm to do classification, regression, and decision making without explicit programming. DL is well-suited for the non-linear data patterns\cite{ref41} and it works in a supervised, partially-supervised, unsupervised or reinforcement manner. Recently, the use of DL is widely observed in transportation networks \cite{DLits}. Different variants of the deep architectures are present in the literature to solve different security problems in vehicular networks. \par 
An Artificial Neural Network (ANN) is from one of the initial designs of neural networks \cite{ANN}, whereas multilayer perceptron (MLP) is the simplest and most-commonly used version of an ANN \cite{MLP}. An MLP is also known as feedforward-ANN. An MLP is a layered model which consists of three sections namely, input layer, middle hidden or computation layers, and output layer. MLP can be modeled as a supervised or unsupervised learning process \cite{DLwireless}. MLP faces the problem of low convergence efficiency and high working complexity because of its fully-connected design. Irrespective of these drawbacks, ANN architectures are used for driver ID fingerprinting, attack detection and  intrusion detection applications in VANET \cite{ref12}.\par   
Another variant, deep neural network (DNN) is an ANN with deep (multiple) hidden layers \cite{DNN2}. The advantages of using a deep structure of layers are: 1) To make the design compatible for learning larger datasets. 2) A deep network learns more complex functions compared to a shallow network, and 3) Deep network architectures are useful in achieving higher accuracy \cite{DeepArchi}. On the other hand, deeper networks introduce much higher complexity in terms of processing time and convergence. Based on the optimization function and information flow, it can be further classified into different types. We discuss below three different types of DNN which find applications for security in vehicular networks.
\begin{enumerate}
	\item Fully-Connected DNN (FCDNN) consists of a series of layers where each input is connected to each neuron of the next layer and so on. Due to this diversified connectivity, it is known as a fully-connected network. 
	\item Convolutional Neural Network (CNN) is one of the popular and widely-implemented DNN in the literature which achieves excellent performance for multi-dimensional data. It is called CNN as it uses the mathematical operation of convolution at different layers in its network design \cite{DeepArchi}. The use of supervised as well as unsupervised methods is widely-observed in CNN \cite{CNN}. This makes it promising for the application of intrusion detection systems (IDS) where an unlabeled dataset is used. CNN has an advantage of good training performance which uses fewer parameters due to weight sharing and pooling operations. CNN achieves the best performance with the multi-dimensional data such as speech and image processing. Several works in the literature use CNN architecture for security applications which are discussed in Section \ref{Sec:LearningSolutions}.
	\item Recurrent Neural Network (RNN) is a recursive DNN in which neurons use a feedback looping structure where new output of a neuron depends on the previous output and the current input \cite{RNN}. RNN is capable of storing states over time. There are many applications for which CNN is not good enough, such as understanding temporal information in videos (i.e. sequence of images) or text blocks. RNN is specially designed to process sequence-data where the current data-point shows some relation with the previous one. The Long Short-Term Memory (LSTM) is a popular RNN design and widely used in vehicular networks which generates long sequences of the time-series data such as traffic flows, sensor readings, and vehicle trajectories.
\end{enumerate}

\subsection{Federated Learning}
\begin{figure}[htbp]
	\begin{center}
		\includegraphics[width=2.8in,height=1.3in]{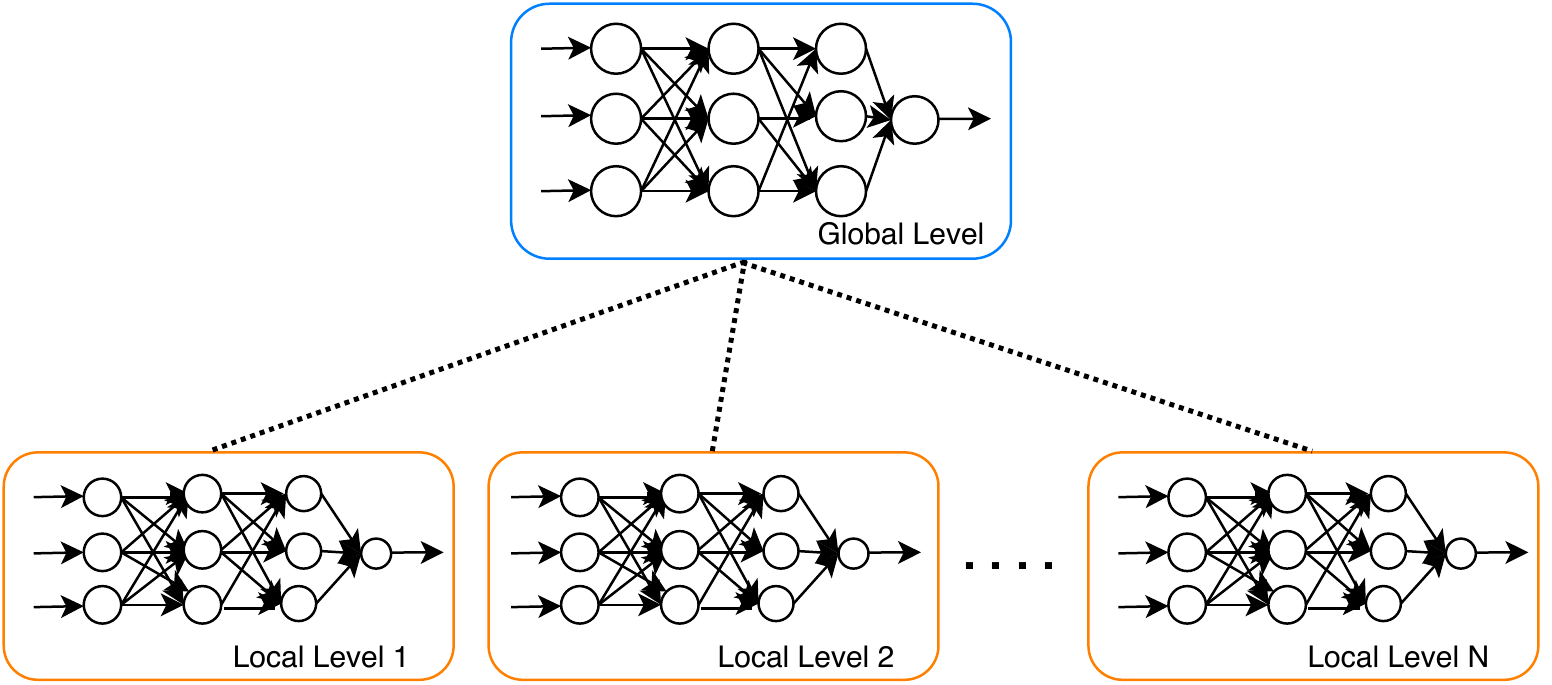}
		\caption{\readded{The conceptual framework of FL.}}
		\label{fig:FDLModel}
	\end{center}
\end{figure}

FL is a recent learning strategy for collaborative training of a series of learning models to reduce communication and processing overhead \cite{FDL}. It works in conjunction with the existing ML models. The FL design follows star topology and works at multiple levels of the network, such as lower-level and upper-level. There can be multiple parallel learning tasks at the lower-level (local-level) and one task at the upper-level (global-level), as shown in Fig. \ref{fig:FDLModel}. The networks at the lower-level are trained based on the obtained weights from the global model to support ensemble learning. Due to its distributed functionality, the FL is also known as distributed learning or lightweight learning. Recently, multiple applications have emerged for vehicular networks that use FL \cite{FDL_TCPPerformIoV}\cite{FDL_latency}\cite{FDL}\cite{FDL2}.

\subsection{Transfer Learning}
\begin{figure}[htbp]
	\begin{center}
		\includegraphics[width=2.5in,height=1.6in]{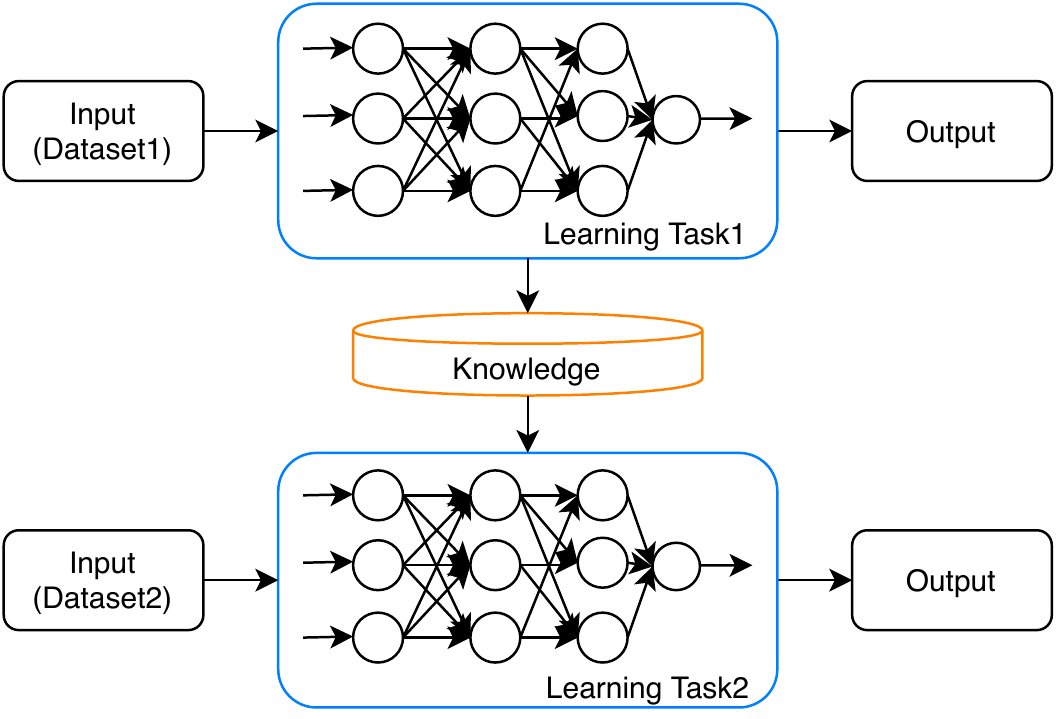}
		\caption{\readded{The conceptual framework of TL.}}
		\label{fig:TLModel}
	\end{center}
\end{figure}

TL is a new type of widely used learning strategy\cite{TL}. As the name suggests, it transfers important parameters among networks to fine-tune its working mechanism for faster and efficient performance. In DL, a network goes through a process of training to learn the optimal weights and bias values. The idea of TL is to use the output of one network for the training of another network, as shown in Fig. \ref{fig:TLModel}. The weights and bias values are learned from one network and directly applied for the optimal working of another network instead of going through the whole process of training again \cite{TL2}. This helps to accelerate the network performance by reducing the training time for a network. Different paradigms of TL are proposed in the literature including, one-shot learning, deep one-shot learning, zero-shot learning, etc, for different applications \cite{DLwireless}. TL finds useful application in mobile networks where state changes occur frequently. In the context of vehicular network security, TL is a competent strategy where new attacks are detected based on the transferred knowledge of the old attacks.

\section{ML-based Security Solutions for Vehicular Networks}
\label{Sec:LearningSolutions}
\begin{figure*}[htbp]
	\begin{center}
		\begin{forest}
			for tree = {
				draw,
				grow=0,
				edge={->,>=latex},
				s sep=2mm, 
				inner sep=4, 	
				anchor=parent,
				forked edges,
			}
			[\readded{Security Solutions}
			[\readded{Privacy Protection} [\readded{\cite{ref79, ref80,PPFL,FLSecureData,FLPP2}}]]
			[\readded{Trust Computation} [\readded{\cite{ref69, ref61, ref68, ref66, ref67, ref70, ref81,nRef5,nRef6,nRef7}}]]
			[\readded{Misbehaviour or Intrusion Detection} [\readded{\cite{ref51, ref47, ref45, ref58, ref59, ref39, ref48, ref34, ref43, ref31,nRef1,nRef2,nRef4,nRef8,TLIDS,CANTLIDS,SupMDS,DLIDS}}]]
			[\readded{Attack Detection} 
			[\readded{Miscellaneous} [\readded{\cite{ref24, ref53,RLCPAttack,DLCPAttack}}]]
			[\readded{Spoofing Attack} [\readded{\cite{ref57, ref54, ref55}}]]
			[\readded{Jamming Attack} [\readded{\cite{ref38, ref50, ref36, ref40}}]]
			[\readded{Sybil Attack} [\readded{\cite{ref30, ref27, ref33}}]]
			[\readded{Black-Hole and Grey-Hole Attack} [\readded{\cite{ref32, ref26}}]]
			[\readded{DDoS} [\readded{\cite{ref25, ref46,RLDOS,USAttack}}]]
			[\readded{Platoon Attack} [\readded{\cite{ref22}}]]
			]
			[\readded{Driver Identification/Fingerprinting}
			[\readded{\cite{ref8,ref10, ref9, ref11, ref12, ref13, ref14, ref15, ref16, ref17, ref18, ref19, ref20, ref21, Ref5,nRef3}}]
			]
			]
		\end{forest}
		\caption{Security solutions and ML-based literature.}
		\label{fig:LearningSolutions}
	\end{center}
\end{figure*}
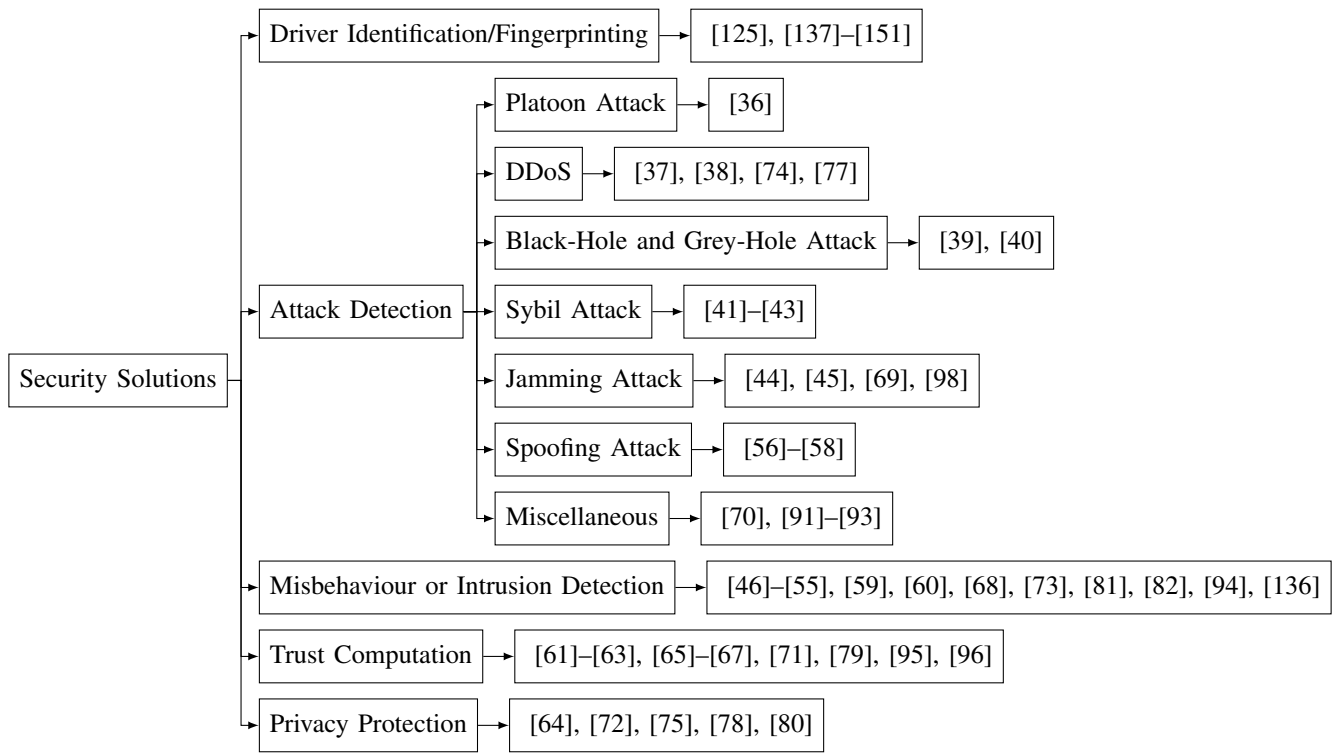
In this section, we discuss different ML-based security solutions for vehicular networks proposed in the literature. Fig. \ref{fig:LearningSolutions} presents a classification of security solutions where the use of ML is widely exploited in the literature to protect vehicular nodes and related information. The description of ML techniques targeted security area, and its usage in vehicular networks are briefly explained in the following sections.

\subsection{Driver Identification/Fingerprinting}
In vehicular systems, data generated from sensors involve information about the vehicle and driver. The driver connects to the vehicle in multiple ways including, personal phone, vehicular services, infotainment services, insurance details, manufacturer services, camera sensors, online accounts, and so on \cite{Ref2} \cite{Ref1}. It is of utmost importance to preserve the privacy of driver data, maintain security and prevent adversaries from tracking the vehicle by linking it to the driver's identification. Different techniques are available in the literature and most of them are based on the idea of hiding location and accessing resources/services without disclosing the true identity (i.e, using pseudonymity) of a driver to ensure privacy \cite{Ref7} \cite{Ref3}. \par

The autonomous vehicles should maintain correct identification and true profiling of the driver to prevent hacking and theft of the vehicle. To enable this, ML techniques are used by researchers to authenticate the true identification and fingerprinting of a user/driver to make vehicles resilience to the theft without hiding the true identity of drivers. Recently, there has been an immense investigation on the driver behavioural data and driving patterns to identify the true user. The data from smartphone sensors and on-board electronic control units (i.e. using the on-board diagnostic (OBD)-II protocol) are most-commonly used for driver profiling. Few studies in the literature focus on biometric and key-based authentication techniques for driver identification \cite{Biometric1,Biometric2,KeyAuth}. However, these techniques are difficult to achieve higher accuracy and implement in the low-powered vehicle security systems. Therefore, recent works deal with driver identification by applying ML algorithms to behavioural data. The summarized tabular comparison of solutions using ML for driver fingerprinting is presented in Table. \ref{tab:driverID}. \par

\begin{table*}[htbp]	
	\centering	
	\caption{Summary of Literature on Driver Identification or Fingerprintings using ML}	
	\begin{tabular}{ccP{3.5cm}P{3.5cm}P{3.5cm}}		
		\toprule
		\textbf{Citation} & \textbf{Year} & \textbf{Feature Data} & \textbf{Algorithm} & \textbf{Accuracy} \\		
		\midrule
		\cite{ref8} & 2012  & Brake pedal pressure, gas pedal pressure, vehicle velocity and the distance from the vehicle in front signals & Temporal clustering with HMM & 70\% (group of 23), 85\% (group of 3) \\		
		\cite{ref10} & 2013  & Acceleration, braking and turning events & SVM and K-mean clustering & 65\% (group of 2/3) \\
		
		\cite{ref9} & 2015  & 21 features including speed, distance, direction and pressure & ELM & 75\% (group of 11), 86\% (group of 5), 88\% (group of 4), 90\% (group of 3) \\
		
		\cite{ref11} & 2016  & 16 features from OBD-II  & Random Forest, Naive Bayes, KNN and SVM & 87\% accuracy (group of 15, few sensors), 99\% (group of 15, more sensors) \\
		
		\cite{ref12} & 2016  & 15 features from OBD-II  & Decision Tree, Random Forest, KNN and MLP & 99\% (group of 10) \\
		
		\cite{ref13} & 2016  & GPS data & DL & 77\% (50 drivers), 60\% (1000 drivers) \\
		
		\cite{ref14} & 2017  & Trip-based data & SVM, Random Forest and Naive Bayes & 88\% \\
		
		\cite{ref15} & 2017  & Repulsive potential energy of vehicles & SVM   & 70\% (group of 4) \\
		
		\cite{ref16} & 2017  & Acceleration & NN & 88\% (group of 13) \\
		
		\cite{ref17} & 2017  & Acceleration and decelration & Multiclass LDA & 60\% (group of 5) \\
		
		\cite{ref18} & 2017  & GPS data & Autoencoder regularized DNN & 78\% (group of 50) \\
		
		\cite{ref19} & 2018  & 137 statistical features extracted from GPS data & Random Forest  & 82\% (group of 3/4) \\
		
		\cite{ref20} & 2018  & 51 features from OBD-II & J48, Random forest and REPtree & 99\% (group of 10) \\
		
		\cite{ref21} & 2018  & 10 sensor readings from OBD-II & DL & 90\% (group of 4) \\
		
		\cite{Ref5} & 2019  & Smart phone sensor + OBD-II protocol data & CNN and RNN & 95\% (group of 10, 0 anomalies), 57\% (group of 10, 50\% anomalies) \\
		
		\cite{nRef3} & 2020  & The intake air pressure and torque of friction & ML (10 schemes compared) & 90\%-93\% \\
		
		\bottomrule
	\end{tabular}%
	
	\label{tab:driverID}%
	
\end{table*}%

In \cite{ref8}, the authors present the idea of temporal clustering with a hidden Markov model and applying mean-square error to each cluster for predicting driver behaviour. The clusters are created based on the actions of the driver in terms of brake pedal pressure, gas pedal pressure, velocity, and the distance it keeps with the front signal. The distraction due to the road conditions shows a great impact on these features. This work achieves an accuracy of about 70\% among 23 drivers and 85\% among 3 drivers when the road distraction is considered. \par 
In the literature, solutions have been developed using SVM, k-means clustering, random forest, naive Bayes, and KNN algorithms for driver profiling and true identification. In \cite{ref10}, the authors demonstrate the potential of using inertial sensors to identify the driver. They consider turning events, braking, and acceleration to differentiate among the drivers. The results are obtained by applying SVM and K-means clustering to the inertial sensors data and accuracy of 65\% is achieved to differentiate the identity among 2 or 3 drivers. In \cite{ref9}, the authors consider 21-unique features to propose a highly-accurate personalized driver assistance system. The features include speed, distance, direction, pressure, and so on, of different objects in a vehicle. A new ML technique called Extreme Learning Machine (ELM) is used in this work which achieves the accuracy rate of 75\% for a group of 11 drivers, 88\% for a group of 5 drivers, and 90\% for a group of 3 drivers. Similar to \cite{ref9}, the authors in \cite{ref11} and \cite{ref12} also consider the use of a wide range of features i.e. 16  and 15 different measurements from OBD-II protocol, respectively. In \cite{ref11}, the authors compare 4 different ML algorithms (i.e. random forest, naive Bayes, KNN, and SVM) to perform authentication with an accuracy of 99\% among 15 drivers. However, the work in \cite{ref12} achieves an accuracy of 99\%  among 10 drivers by using a decision tree, random forest, KNN, and MLP algorithms. \par 
The work in \cite{ref14} highlights different kinds of features for driver identification where the authors collect trip-based data to help authenticate a driver. The traditional ML algorithms including SVM, naive Bayes, and random forest are used to achieve an accuracy rate of 88\%. Another unique set of features is considered by the authors in \cite{ref15} to profile a driving style. In this work, the repulsive potential energy values generated from the preceding vehicles have been used. A simulator collects data for training and testing purposes. Using the SVM classification method, this work achieves 70\% accuracy among 4 drivers. In \cite{ref20}, a different kind of decision tree-based ML technique is applied over 51 features of the OBD-II protocol data. Such a large number of features helps to obtain about 99\% accuracy among 10 drivers. Another work in \cite{ref19} uses GPS data to perform driver identification with the random forest algorithm by extracting 137-additional statistical features to achieve an accuracy of 82\%. \par 
In recent years, there have been research that focus more on achieving high accuracy with a minimum number of features and/or automatic feature extraction. It is difficult to handcraft a large set of features in order to achieve a higher identification score. The effect of distractions from the road or location is also challenging for the manually-chosen type of features. This has lead researchers to explore the use of DL to let the algorithm automatically extract features from the sensor data. In this context, the first work on DL for driver identification is proposed in \cite{ref13}. The authors in this work use GPS data as input to achieve 77\% accuracy among 50 drivers and 60\% accuracy among 1000 drivers. Another work in \cite{ref18} presents the idea of feature extraction (from GPS data) by using autoencoder-regularized DNN with an accuracy of 78.3\% among 50 drivers. In \cite{ref16} and \cite{ref17}, the acceleration and deceleration data are considered for driver identification and classification. The authors in \cite{ref16} train and test their design by applying an NN over 4 different clusters of data obtained from 13 drivers. They are able to achieve an accuracy of 88\% for the correct identification of drivers. In \cite{ref17}, the authors propose a multiclass-LDA classifier to identify drivers. Their simplest design obtains an accuracy of 60\% among 5 drivers. \par 
An analytical study is performed in a recent work \cite{nRef3} over ten different ML algorithms for accurate detection of the true driver. This work aims to use minimum number of features for driver prediction which includes intake air pressure and torque of friction. In the performance evaluation, an accuracy of 93\% is achieved by multiple ML techniques to discriminate between true driver and impostor. The model discussed in \cite{ref21} inputs 10 different sensor readings of OBD-II protocol to the CNN. This work presents an 8-layer CNN design which is able to achieve an accuracy of 90\% for 4 drivers within a few minutes. In a recent work \cite{Ref5}, the authors employ behavioural data of drivers with a DL algorithm to guarantee driver identification while considering the effect of anomalies. They propose an end-to-end driver fingerprinting framework by using CNN and RNN over the combined data of smartphone sensors and the vehicle's electronic control unit. In this work, the authors consider a driver's personal data as a time-series sequence and perform identification through the multivariate time-series classification. It achieves an accuracy of 95\% (no anomalies) and 57\% (50\% of anomalies) for 10 drivers. \par

\subsection{Attack Detection}
\label{Sec:attackDetection}
Vehicular networks are vulnerable to different types of attacks and a number of solutions are proposed in the literature to deal with such attacks\cite{ref42}. The evolution of V2R, V2V, and V2I communications demands highly-efficient, intelligent, and faster solutions. The traditional hard-coded algorithms are able to deal with the deterministic type of attack scenarios. On the other hand, the self-learning designs of deep architectures can detect a variety of attacks through experience and shared information. Therefore, ML and its sub-classified architectures are gaining popularity to detect attacks and deal with different types of security issues in V2X communications. In this section, we will discuss different types of attacks and ML-based solutions proposed in the literature to prevent these attacks. The summarized tabular comparison of solutions using ML for attack detection is presented Table. \ref{tab:attack}. \par

\begin{table*}[htbp]

	\centering

	\caption{Summary of Literature on Attack Detection using ML}

	\begin{tabular}{cP{1cm}P{3.5cm}P{3.5cm}P{2.5cm}P{2cm}P{2cm}}

		\toprule

		\multicolumn{1}{c}{\textbf{Citation}} & \multicolumn{1}{c}{\textbf{Year}} & \multicolumn{1}{c}{\textbf{Feature Data}} & \multicolumn{1}{c}{\textbf{Algorithm}} & \multicolumn{1}{c}{\textbf{Type of Attack}} & \multicolumn{1}{c}{\textbf{Service}} & \multicolumn{1}{c}{\textbf{Accuracy}} \\

		\midrule

		\cite{ref32} & 2015  & Auditable data from basic, IP and AODV trace files & Artificial NN & Black hole & Detection & 99\% \\

		\cite{ref26} & 2016  & Rx packets, PDR, dropped packets and delay & NN and SVM & Grey hole and rushing attack & Detection & 99\% (both) \\

		\cite{ref30} & 2017  & Driving pattern & KNN & Sybil attack & Detection & 100.00\% \\

		\cite{ref27} & 2017  & Driving pattern & SVM   & Sybil attack & Detection & 92\%-98\% \\

		\cite{ref25} & 2018  & 7 features extracted from TCP and UDP flows & ML (8 schemes compared) & DDoS  & Detection & NA \\

		\cite{ref38} & 2018  & History of power usage & Dyna Q-RL & Jamming attack & Prevention & NA \\

		\cite{ref24} & 2018  & Acceleration by LIDAR and RADAR & HMM   & Attack on cruise control algorihm & Detection & 78\% to 90\% \\

		\cite{ref46} & 2018  & OpenFlow flow tables & SVM   & DDoS  & Detection & 98\% \\

		\cite{ref50} & 2018  & RSSI, PDR, SINR and RSV & K-means clustering & Jamming attack & Detection & NA \\

		\cite{ref57} & 2018  & Physical layer information & RL (Q-Learning) & Spoofing attack & Detection & NA \\
		
		\cite{RLCPAttack} & 2018 & Sensor readings and beaconing & DL and RL & Cyber physical attack & Prevention & NA \\

		\cite{ref22} & 2019  & Range and veloctiy by LIDAR and RADAR & CNN and FCDNN & Attack to destablize the platoon & Detection & 98\% \\

		\cite{ref33} & 2019  & 11 features including range, position, speed etc & RNN (LSTM) & Sybil attack & Detection & 95\% \\

		\cite{ref36} & 2019  & History of actions and observations & RL (Q-Learning) & Jamming attack & Prevention & NA \\

		\cite{ref40} & 2019  & RSSI, ToA, ToD and distance & CatBoost (Decision tree ML) & Jamming attack & Detection and localization & 100\% \\
		
		\cite{RLDOS} & 2019  & Packet features from live network traffic & RL (Q-Learning) & DDoS attack & Detection & NA \\

		\cite{ref53} & 2019  & SDN traffic flows & ANN, DL and LSTM & Crossfire attack & Detection & 80\%-87\% \\

		\cite{ref54} & 2019  & RSS, RSU location and spoofed location & Multi-layer NN & Spoofing attack & Detection & NA \\

		\cite{ref55} & 2020  & Spatial decorrelation features & RL (Q-Learning) and TL & Spoofing attack & Prevention & NA \\
		
		\cite{USAttack} & 2020 & Car-hacking dataset & K-means & DoS, fuzzy, RPM and gear attack & Detection & 99\% \\
		
		\cite{DLCPAttack} & 2020 & Sensor readings and beaconing & DL and RL & Data manipulation attack & Prevention & NA \\

		\bottomrule

	\end{tabular}%

	\label{tab:attack}%

\end{table*}%

\subsubsection{Platoon Attack}
Platoon is a concept of grouping vehicles which travel in the same lane with close proximity and similar speed regulations \cite{ref23}. Platoon helps to save energy and improves road capacity to efficiently manage traffic. However, attacks on the stability of a platoon may result in car accidents and severe losses. The authors in \cite{ref22} propose a self-learning deep architecture to detect attacks against the stability of a platoon. A platoon of 10 vehicles is considered where the first vehicle is a leader vehicle with the desired speed and all other vehicles in a platoon follow it. The attacker tries to destabilize a platoon by applying constant brakes or abrupt accelerations to deviate the speed of nearby vehicles. In \cite{ref22}, by using FCDNN and CNN, the authors are able to detect and locate the attacker. The velocity, range, and distance obtained from LIDAR and RADAR sensors (present in all autonomous vehicles) are fed into the FCDNN and CNN networks. The last hidden layer of the network contains 10 outputs corresponding to each of the vehicles in a platoon. A high (or 1) at any output position indicates the presence of attack and its location. This design is able to achieve an accuracy of 97\% in detecting the attacks.

\subsubsection{DDoS}
The use of SDN is gaining a lot of importance in the field of transportation systems. However, the centralized intelligence of SDN poses some threats. It provides an easy point of access to launch distributed DoS (DDoS) attacks by generating a large number of spoofed-flows requesting SDN services. The authors in \cite{ref25} use an ML-based solution to detect DDoS attacks in SDN-based vehicular networks with a focus on V2I communication. It performs an analytical study to find the best ML mechanism to detect DDoS attacks for the given conditions. It collects ground truth data from transport control protocol (TCP) and user datagram protocol (UDP) flows (as 7 different features) in the presence as well as an absence of DDoS attacks. On the basis of collected data, it trains the network with multiple supervised learning algorithms and observed that the gradient boost classifier achieves the best performance. In addition, the random forest, decision tree, and linear SVM techniques perform close to the best performance. However, the NN does not perform well because of the insufficient amount of data used to train the NN. \par 

The authors in \cite{ref46} study DDoS attacks based on TCP flood, UDP flood, or ICMP (internet control message protocol) flood in an SDN-based vehicular network. In this work, the authors use the SVM algorithm to identify efficiently and respond quickly in the event of an attack. The flow table entries are used as input features for the SVM training. The data forwarding plane forwards every new entry towards the controller. Over the control plane, the PACKET\_IN trigger check takes place and the rate of PACKET\_IN message is compared with the threshold rate. In case of an abnormality, a warning message is transmitted to the attack detection module where the SVM recognition algorithm checks the presence of an attack by extracting the flow characteristics. If confirmed, the attack warning system generates alert proceeds with further actions. The authors evaluate the design in terms of detection ratio, false alarm ratio, and classification time. An average detection ratio of $\approx$98\% is achieved for different types of floods with the longest classification time of 0.148sec. \par
Sherazi et. al \cite{RLDOS} propose an idea of analyzing packets from live network traffic to detect vulnerabilities. This work focuses on DDoS attack detection using RL in IoV networks. This work uses cluster-based topology where cluster heads collect sensor data from end-users. The proposed algorithm is designed to perform attack detection on live traffic where it captures and analyzes packets using fuzzy logic. Further, it uses Q-learning over pre-processed packets to detect DDoS attacks. In this work, the authors use NS-3 simulation platform to evaluate the proposed method using buffer size usage, energy consumption, response time, and throughput as performance metrics. Angelo et. al \cite{USAttack} propose a data-driven approach to detect DoS attacks along with three other types of attacks on in-vehicle networks. This work uses data associated with the controller area network (CAN) bus and extracts useful features using unsupervised learning. It represents CAN behaviour from those features and any deviation from the learnt behaviour is considered as an attack on the system. Further, this work uses a data-centric scheme to narrow down the type of an attack by observing associated parameters of the CAN bus. The performance evaluation results of the proposed scheme show an accuracy of 99\%-100\% for a car-hacking dataset.

\subsubsection{Black-Hole and Grey-Hole}
The black-hole and grey-hole attacks are types of wireless routing attacks. In these attacks, a node tries to stop onward forwarding of messages/packets towards the receiver. In a black-hole attack, there will be a complete blackout or drop of packets. On the other hand, in a grey-hole attack, packets are selectively dropped and a subset of packets are altered by an attacker to convey wrong information to the receiver. In the literature, ML techniques have been developed to detect these kinds of routing attacks.\par 
In \cite{ref26}, the authors present a design to detect grey-hole and rushing attacks using NN and SVM algorithms. The purpose of both attacks is to disconnect links between RSU and vehicles which prevents the discovery of routes for the packet transfer. In this work, the authors record the number of received packets, dropped packets, PDR, and average end-to-end delay for normal and malicious behaviour in VANETs. The recorded data is pre-processed, trained, and tested with the feed-forward NN to collect output in terms of normal and abnormal behaviour. Further, the same data is trained and tested with the SVM model as well to identify system efficiency for detecting grey-hole attacks and rushing vehicles. An alarm is generated to notify systems regarding the attacks based on a joint-decision from both techniques. As a result, the above models achieve a maximum error rate of 0.19\% and 0.17\% for SVM and NN, respectively. \par 
Alheeti et al. \cite{ref32} propose an ANN design to detect black-hole attacks in self-driving cars. This work considers three different types of trace files namely, basic trace file, IP trace file, and AODV trace file, to extract features for the network. These files generate a large number of features. To find the most effective features, the "proportional overlapping scores" method is used in this work. The authors use 21 features to characterize data as normal and abnormal. This data is used as input for the NN to detect the black-hole attacks. Using the NS2 (network simulation-2) simulator, the authors demonstrate accurate attack detection with a rate of 99.8\%.

\subsubsection{Sybil Attack}
Sybil attack is one of the most common and easy-to-implement attacks. A sybil attack in a vehicular network creates virtual nodes to launch an attack and detection of such virtual nodes is not easy. The use of pseudonyms in vehicular networks for the privacy of user identity makes it difficult for the system to detect sybil attacks. \par 
Different techniques are proposed in the literature to enhance the authentication mechanism and prevent unauthorized access \cite{ref28}\cite{ref29}. Pengwenlong et al. in \cite{ref27} apply an ML algorithm over the driving patterns to successfully detect sybil attacks in vehicular networks. The idea is to find the similarity between driving patterns based on time, location, velocity, acceleration, and acceleration-variation in order to detect malicious nodes. A new concept of driving pattern matrix (DPM) is proposed in this work which uses eigen values as input to the SVM algorithm. The results are obtained by varying different parameters and using different SVM kernels to achieve an accuracy of 92\% to 98\%. Another work by Pengwenlong et al.\cite{ref30} is also about sybil attack detection but using the KNN technique. In this work, the authors use the driving pattern and eigen values of DPM to detect an attacker with an accurate classification rate of 100\%. \par 
Kamel et al. \cite{ref33} present a generic RNN-based solution to perform global detection of sybil attacks. This work considers four different effects of a sybil attack which results in traffic congestion, data replay, DoS random, and DoS disruptive. In the first step, OBUs and RSUs detect the misbehaviour within vehicles. In the case of malicious activity, it reports to the misbehaviour authority (MA) by sending misbehaviour reports (MBRs). The function of MA is to achieve a global view of misconduct that takes place in vehicular nodes. MA performs 11 different checks of information including, range plausibility, position plausibility, speed plausibility, and so on, to use it as an input for an LSTM-based RNN network which detects the correct type of a sybil attack. In this work, the authors also employ feature compression by using an autoencoder algorithm. They evaluate the performance of their proposed model using OMNET++ simulator which shown an accuracy of 95\%.

\subsubsection{Jamming Attack}
In this attack, the aim of a jammer is to block or interfere transmission of data from a sensor by sending false alerts or creating a spoofed environment around the sensor. The trust-based mechanisms are commonly used to prevent access of jammers to the system but sometimes these techniques fail to differentiate between the trusted node and malicious node. Xu et al. \cite{ref36} present an RL-based jamming attack prevention algorithm for mobile ad-hoc networks. The proposed algorithm performs Q-learning to learn the history of actions (past), and then input it into the deep Q-network (DQN) to predict Q-values for the present states. It is observed from the performance results that a well-learned network (based on past experiences) helps the transmitter to perform optimally and quickly in case of a jamming attack. The performance evaluation of the proposed mechanism is carried out using tensorflow\cite{ref37}. \par 
A jamming attack against a vehicle platoon is discussed in \cite{ref38}. This work presents a Dyna-Q RL-based power control mechanism. In this scenario, the jammer tries to consume channel energy and prevents an efficient transmission of data among vehicles. The RL-based algorithm performs a historic assessment of the environment to calculate the mean approximation of a channel utility and uses it to control the power for vehicles. The proposed mechanism not only prevents jamming but also enhances the signal-to-interference noise ratio (SINR). \par 
Kumar et. al \cite{ref40} present an anti-jamming protocol to detect and localize the jammers in vehicular networks. This work contains multiple stages of functionalities in which at first a foster rationalizer is implemented to identify the frequency change that results in signal strength variations. The next step is to use a morsel supple filter for the minimization of noise to perform accurate localization. A decision tree-based ML algorithm (known as CatBoost) is used to locate a jammer. It employs 4 different features for the correct prediction of location i.e. distance factor, time of arrival (ToA), time of delay (ToD), and received signal strength indicator (RSSI). In their performance evaluation, the proposed algorithm maintains accurate location prediction accuracy of 99.91\% using MATLAB/SIMULINK. The proposed algorithm also achieves high throughput, high PDR, and low packet loss ratio. \par 
Karagiannis et al. \cite{ref50} propose an unsupervised algorithm to detect RF-jamming in vehicular communications. In this work, three real-time simulations are evaluated to identify intentional and unintentional jamming of RF signals in V2V communications. It implements RSSI, PDR, SINR, and relative speed variation (RSV) from the vehicle's OBU as input features to an unsupervised classifier. The authors claim the RSV parameter is a key factor to differentiate between jamming due to malicious nodes and jamming due to unintentional system problems.  

\subsubsection{Spoofing Attack}
In vehicular applications, different types of spoofing attacks can take place wherein an attacker pretends as a legitimate user of the network to gain access to the personal information. A spoofing attack is not only limited to the spoofing of identification but also the location, DNS information, IP address, and so on. In \cite{ref54}, Ihsan et al. propose a model to encounter the problem of location spoofing in vehicular networks. It considers traditional location verification systems (LVS) as inefficient due to their relevance on channel parameters which limits its application for the highly-mobile real-time scenarios in vehicular networks. This work proposes a multi-layer NN to classify node location as legitimate or spoofed. The received signal strength, untrue (spoofed) locations, and RSU locations are feed into the network to generate binary responses. As an evaluation parameter, the error function is used to compare the proposed work with the existing traditional methods to verify the effectiveness of using ML in identifying spoofed entities. This work considers scenarios where it is believed that a malicious node is optimizing its untrue location to make it hard for a network to detect an attack. However, ML-based NN design works efficiently to detect an attack for the optimized scenarios as well. \par 
Lu et al. \cite{ref57} present an RL-based spoofing attack detection mechanism. In this work, the authors implement a Q-learning authentication mechanism to detect rogue nodes. This model uses physical layer information such as RSSI to find spoofing data by sending an alarm when there is a mismatch between the information received from a rogue node and the previously recorded data (of old legal users) in a network. It improves the detection accuracy by maintaining a record of radio sources and physical layer parameters. The final decision is made by Q-learning which is trained by performing repeated spoofing detections in the form of a Markov decision process. Prior knowledge of the attack model and network model is not important in this kind of architecture. In their performance evaluation, the proposed model achieves lower misdetection and false alarm rate compared to the existing solutions. \par 
The authors of \cite{ref55} extend \cite{ref57} with an advanced and robust RL-based physical authentication mechanism to prevent unknown spoofing attacks in VANETs. The existing physical authentication schemes use channel state information to prevent an attack \cite{ref57}\cite{ref56}. In the extended work, the authors propose a low-energy-consuming mechanism which performs authentication without the prior knowledge of channel parameters. The proposed work performs RL to detect spoofed packets based on physical spatial decorrelation features through a trial-and-error method. The authentication policy is selected based on the current state and its Q-value to authenticate packets with the user identity. This work employs TL as well to save convergence time and make the learning process faster. The results are compared with earlier RL-based work to prove the better performance of their proposed authentication scheme.  

\subsubsection{Miscellaneous}
In addition to the above well-known attacks, there is a possibility of other types of attacks that can take place at different parts of vehicles and vehicular networks to affect the normal working mechanism. Jagielski et al. \cite{ref24} propose an ML-based mechanism to detect four different types of attacks (i.e. ACL (Acceleration), VEL (Velocity), POS (Position), and VEL-POS) which can compromise the cruise control system of a vehicle. Here, ACL and VEL can impact the passenger comfort and efficiency, respectively. However, POS and VEL-POS can result in a car crash. The proposed model performs attack detection by using a physical kinematics equation and a hidden Markov model. In this work, the information regarding acceleration collected from LIDAR and RADAR sensors is used as an input feature for the ML model. In their performance evaluation, the authors detect the above mentioned four types of attacks with an accuracy rate of 78\% to 90\%. \par 
Narayanadoss et. al \cite{ref53} present an ML-based solution to detect crossfire attacks where an attacker disconnects the set of links/nodes from the rest of the network. An SDN-based intelligent transportation system is considered in this work. It performs a comparison analysis between ANN, DL, and LSTM to find the highly-accurate attack detection model. In the SDN-based design of a vehicular network, the traffic behaviour and temporal correlation between the traffic flows are recorded to use as an input for the ML models and differentiate between the legitimate and attacker flows. In this work, the authors use mininet platform to perform a comparative analysis between different ML-models in vehicular network scenarios. The results conclude that LSTM outperforms ANN and DL in accurately detecting the attacker flows based on different parameters. The average detection accuracy achieved by ANN and DL is about 80\% and for LSTM it is about 87\%. 
Ferdowsi et. al \cite{RLCPAttack} propose a method to increase the robustness of an autonomous vehicle's dynamics control system in case of cyber-physical (CP) attacks. Such CP attacks are a type of attacks on vehicles where adversaries try to manipulate their sensor or communication data. In this work, the authors use a game-theoretic approach to formulate an action where the attacker injects faulty data into the control system to manipulate its optimal safe spacing measurements. In the reaction, the task of the vehicle's defending system is to maximize the robustness against faulty data. The attacker can inject faulty data using an infinite range of data values and the defending system has no information for the attacker. In the proposed work, each player of the vehicle's system uses an LSTM network to learn its own action and feeds it into the deep RL algorithm. The RL algorithm trains itself with actions that minimize the spacing deviation. Any value which results in maximum deviation would be considered as an attack. The performance evaluation shows deep RL helps to prevent the CP attacks and makes the dynamics control system robust against illegitimate changes. Another recent work, with the same application of enhancing the robustness of autonomous vehicle's dynamics control system, is proposed in \cite{DLCPAttack}. This work also uses game theory along with LSTM and deep RL. The difference from the previous work is the use of additional generative adversarial network (GAN) architecture to enhance the robustness and prevent control system data manipulation.

\subsection{Misbehaviour or Intrusion Detection}
\label{Sec:MDS}
\begin{table*}[htbp]
  \centering
  \caption{Summary of Literature on Misbehaviour or Intrusion Detection using ML}
    \begin{tabular}{cP{1cm}P{3.5cm}P{3.5cm}P{2.5cm}P{2cm}P{2cm}}
    \toprule
    \multicolumn{1}{c}{\textbf{Citation}} & \multicolumn{1}{c}{\textbf{Year}} & \multicolumn{1}{c}{\textbf{Feature Data}} & \multicolumn{1}{c}{\textbf{Algorithm}} & \multicolumn{1}{P{2.5cm}}{\textbf{Type}} & \multicolumn{1}{c}{\textbf{Service}} & \textbf{Accuracy} \\
    \midrule
    \cite{ref51} & 2011  & Wide range of features related to packet delivery, speed, positioning etc & ML (5 schemes compared) & MDS   & Detection & 92\% (single), 93\% (multi) \\
    \cite{ref47} & 2015  & Behavioural and contextual information & SVM   & IDS   & Detection & NA \\
    \cite{ref45} & 2015  & Road traffic and network data parameters & K-OCSVM & IDS   & Detection & NA \\
    \cite{ref58} & 2015  & PDR, MDR, SSI and packet sent &  SVM  & IDS   & Detection & NA \\
    \cite{ref59} & 2016  & Vehicle behaviour (Packet transmission) & SVM   & IDS   & Detection & NA \\
    \cite{ref39} & 2016  & Speed and speed error & NN    & MDS & Detection and Prevention & NA \\
    \cite{ref48} & 2017  & KDD CUP IDS dataset & Multi-class SVM & IDS   & Detection & ~88\% \\
    \cite{ref34} & 2017  & NGSIM dataset & ANN   & MDS & Detection & 99\% \\
    \cite{ref43} & 2018  & VeReMi dataset & SVM and KNN & MDS & Detection and Localization & NA \\
    \cite{DLIDS} & 2018 & Network incoming and outgoing, CPU, disk data, encoder, accelerometer, power and current & RNN & IDS & Detection & 90\% (known attack), 67\% (unknown attack) \\
    \cite{ref31} & 2019  & Beacon & ML (5 schemes compared) & MDS & Detection & 95\% \\
    \cite{SupMDS} & 2019 & VeReMi dataset & KNN and SVM & MDS & Detection & 99\% \\
    \cite{nRef2} & 2020  & CIC-IDS2017 dataset & Feed-forward NN & IDS & Detection & 99\% \\
    \cite{nRef4} & 2020  & NSL-KDD dataset & Ensemble Learning & IDS & Detection & 97\% \\
    \cite{nRef8} & 2020  & Alert-specific features & ML (5 schemes compared) & MDS & Detection & 96\%-98\%\\
    \cite{CANTLIDS} & 2020 & CAN dataset & TL & IDS & Detection & 88\%-95\% \\
    \cite{TLIDS} & 2021 & AWID dataset & TL & IDS & Detection & 92\%-96\% \\
    \cite{MLDCMDS} & 2021 & VeReMi dataset & ML (6 schemes compared) & MDS & Detection & NA \\
    \bottomrule
    \end{tabular}%
  \label{tab:MDS}%
\end{table*}%

Misbehaviour detection system (MDS), also known as IDS is a means of detecting an unknown type of attacks. It is crucial to design a system which identifies the misbehaviour that occurs in any form. Several studies have been carried out in the literature on the problem of intrusion or misbehaviour detection. In this section, we discuss ML-based solutions to handle intrusion or misbehaviour taking place by a dishonest node. The summarized tabular view of solutions is shown in Table. \ref{tab:MDS}. \par 

Grover et al. \cite{ref51} discuss a scenario where a misbehaved vehicle launches an attack by generating false alert messages. In this work, the authors propose an ML algorithm to classify the types of misbehaviours in vehicular networks. It uses single-class ML to differentiate between honest and dishonest nodes, in the first place. In the next step, it applies a multi-class ML to detect the type of misbehaviour or attack that can be launched by a dishonest node. There are six different types of attacks considered in this work which can compromise the authenticity or availability of a vehicle in V2V or V2R communications. This work employs a wide range of features which are collected using the NCTUns-5.0 simulator \cite{ref52}. For the performance evaluation section, five different types of ML algorithms are compared where random forest and J-48 algorithm outperform with a true-positive rate of 92\% for single-class and 93\% for multi-class classifications. \par 

Leandros et. al \cite{ref45} present a distributed IDS (DIDS) in vehicular networks. This work analyses the effect of RSU placement, intruder velocity, and density of vehicles over the accuracy and response time. The DIDS uses K-OCSVM which can be deployed over vehicles or on RSUs for detecting the misbehaviour by an intruder. The K-OCSVM is a combination of one-class SVM (OCSVM) and K-means clustering algorithm. The system uses one-class SVM to separate all possible outliers, and then uses it as an input for the k-means clustering where clustering takes place in a recursive manner to differentiate the most severe alerts into a separate cluster. The separated cluster of negative values is communicated to the security center for further processing. \par  

In \cite{ref58}, the authors develop an IDS using a clustering scheme in VANETs. Compared to the previous works, the proposed algorithm claims not to engage any special agent for observing the node behaviour. In addition, the authors use unique features such as mobility of nodes and topology changes in their framework. It takes a worm-hole attack, black-hole attack, sybil attack, selective-forwarding attack, packet-duplication attack, and resource-exhaustion attack into consideration. It is a lightweight framework that performs detection at multiple levels including, a global decision at RSU, global detection at the cluster head, and local detection at cluster members. At the local level, a Bayesian game is applied by a cluster member that takes features such as PDR, message duplication ratio (MDR), packets sent, and signal strength intensity (SSI) to model rules for the normal behaviour. In case of exploitation of rules, an alarm for the malicious behaviour is generated and then forwarded to the cluster head. However, at the global level (i.e. cluster-head), an SVM learning algorithm confirms the malicious node detection by training it with the same features as used at the local level. Finally, the global level decisions (at RSU) compute the trust level for each node, and the nodes with lower trust values are blacklisted. The performance study verifies that this algorithm is fast, lightweight (in terms of overhead), and has low false-positive rates. \par 

A model named CEAP (Collection, Exchange, Analysis, and Propagation) is proposed in \cite{ref59} which deploys an intrusion detection mechanism on top of clustering protocols. It is a multi-decision intelligent mechanism using SVM classification for intrusion detection. In this work, the cluster heads work as watchdogs to analyse the multi-point relay node and monitor the packet transmission for the classification of malicious and true behvaiours. The main contribution of this work is to maintain low computation power, low communication overhead and reduce the storage usage by deploying it over the existing clustering protocols such as QoS-optimized link-state routing \cite{ref60}) to make it efficient for the infrastructure-less vehicular networks. \par 

Li et al. \cite{ref47} present an IDS which uses the SVM algorithm to detect anomalous vehicles. It employs behavioural and contextual information to train the SVM classifier. In contextual information, the velocity, channel status, temperature, wind speed, GPS coordinates, and altitude are taken into account. This algorithm is more resilient to different attack patterns and environmental changes. It is deployed at every node to analyse the neighbouring node behaviour and exchange their information with one another. In this way, each node gets local as well as external information (shared by others). The dempster-shafer theory is used to fuse data at each node. The broader view of the network makes it easier for all nodes to have the same belief of malicious nodes. In the performance evaluation, the parameters such as communication overhead, precision, and recall are calculated and compared with the previous works. \par  

In \cite{ref39}, Sargolzaei et al. highlight the concept of fault detection to prevent different types of attacks. A misbehaviour or fault initialization in a network corresponds to the beginning of an attack. This work presents an NN-based design to detect the falsification (i.e. false data injection (FDI)) that can be a cause for different types of attacks. The platoon scenario is considered in this work and two controllers are used to keep track of speed and distance of a vehicle. In case of any change from the reference values, the controller sends a notification to the decision-making unit which uses a fuzzy logic-based NN to detect the fault and generates a new value for the safe-gap needed to maintain between vehicles to prevent accident scenarios. It exploits current speed and speed error to output the safe distance alerts for all the vehicles in a platoon. This work uses simulation to evaluate the design in terms of speed (input) and distance (output). \par 

In \cite{nRef2}, an end-to-end design of a feed-forward NN is proposed for intrusion detection. The detailed methodology helps to provide misbehaviour detection against a wide range of attacks. The authors use an MLP on selected features of a recent dataset (i.e CIC-IDS2017). The selection of features before optimizing the hyper-parameter is important in achieving good performance. A key contribution of this work is the implementation of the proposed design on a MicroProcessor Unit (MPU) from STMicroelectronics which is used as a smart gateway to ensure a connection between the vehicle and cloud. In performance evaluation, false positive ratio of less than 1\% and 99\% accurate intrusion detection are achieved for a wide range of attacks using the CIC-IDS2017 dataset. \par 
Kim et. al \cite{ref48} propose a cloud-based SDN design for vehicular applications to perform intrusion detection. In this work, the vehicles and RSUs are combined to form a data plane and a group of vehicles combines to form a cloud. The controller on the infrastructure side along with certification authority (CA) is used to perform a control plane task. All vehicles are programmed to send information on packet drop rate, packet modification rate (PMR), request-to-send (RTS) flooding rate, wireless channel status, packet interval, and packet size as input features toward a multi-class SVM which is deployed at the controller side. The SVM performs classification based on input features and identifies if an attack is taking place or not. This work performs a MATLAB-based simulation to evaluate the proposed design. In order to create an attack, the KDD CUP intrusion dataset is used \cite{ref49}. Four different types of attacks are considered in this work, including DoS attack, probing attack, a user-to-root (U2R) attack, and remote-to-local (R2L) attack. The performance results show that the cloud-based SVM classification model is able to achieve an accuracy of 88\%. \par 

In \cite{ref34}, the authors consider an ANN to detect misbehaviour information. There are seven different types of features extracted from the NGSIM (Next Generation SIMulation) dataset \cite{ref35} to help detect anomalies. In this work, the detection phase consists of 4 phases: data acquisition, sharing, analysis, and decision making. In the first phase, only data collection takes place. In the second phase, the vehicles broadcast their mobility information to all reachable network nodes and other vehicles. The rate of broadcast and transmission delay is used in this phase to analyse the behaviour of vehicles. In the third phase, a set of features representing misbehaviour are grouped together. In the final phase, an ANN is used to perform training and testing for anomaly detection. The performance results show that the trained classifier detects misbehavour with an accuracy of about 99\%. A scheme is proposed in \cite{ref43} to predict as well as locate the misbehaviour. This work uses VeReMi dataset\cite{ref44}, which is designed for V2X security testing. It contains a labeled dataset for the normal and attack behaviour. The ML algorithms perform different plausibility checks using SVM and KNN to classify an attack and predict its location. Three different plausibility checks used in this research include location, movement, and quantitative information. This scheme can classify misbehaviour efficiently by maintaining recall within 5\%. \par 

Loukas et. al \cite{DLIDS} propose a lightweight DL-based IDS model for vehicular applications which can also be offloaded to other network devices/vehicles. As a case study, the authors in this work use small-land vehicles to perform offloading of the continuous task of IDS and demonstrate high attack detection accuracy using a deep learning model. The proposed work detects known attacks with 90\% of accuracy and unknown attacks with 67\% of accuracy. First, the detection model uses deep MLP along with RNN to process time-series data of eight input features collected from communication, processing, and physical properties of the vehicle. Later, an LSTM hidden layer is used to learn the temporal context of different attacks. However, for offloading, a mathematical model is proposed where authors use minimum detection latency as an objective to make computation offloading decisions for the DL model, given the processing demands (available resources) and the reliability of the communication channel is satisfied.\par 

Sohan et. al \cite{ref31} present an ML-based framework to identify vehicle misbehaviour using false alert messages and position falsification. The vehicles update each other regarding different happenings including, road conditions, accidents, emergency vehicles, and collision warnings by sending information via a beacon. In this work, the authors track beacons and information deviation from the normal protocol conditions and use it as an input for the ML algorithms. Five different types of ML algorithms are compared in this work wherein decision-tree classifier achieves the highest accuracy of 95\%. The authors in \cite{SupMDS} propose an ML-based solution to maintain the correctness of information exchanged between V2V and V2I. The misbehaviour performed by an attacker to manipulate or inject false information into the communication stream can cause catastrophes or accidents in vehicles. In this work, the authors introduce three features of the n-sequence trajectory to detect misbehaviour with higher accuracy. It uses supervised KNN and SVM classification and compares it to previous work using the same dataset but with different features. In the performance evaluation, 99.7\% of precision is achieved. \par 

Collaborative IDS is studied by Ghaleb et. al in \cite{nRef4} using ensemble learning and shared knowledge of vehicles. In this work, each vehicle creates an ensemble of weighted random-forest classifiers, for which aggregation takes place using a robust voting scheme. Each vehicle trains local IDS classifiers using a random forest algorithm and shares its knowledge on-demand with other vehicles. The performance of the classifier on each vehicle is evaluated by testing the local dataset over the received classifier, and based on the generated trustworthiness factor of the received classifier. The classifiers which highly-deviate the results are excluded from the ensemble of weighted random-forest classifiers. This work uses network security laboratory-knowledge discovery data mining (NSL-KDD) dataset to simulate performance over four different types of attacks. It can classify attacks with an F1 score of 97\% and 4\% false-positive rate. \par 

Gyawali et. al \cite{nRef8} propose an MDS mechanism to prevent internal attacks. This work uses ML along with reputation theory to detect an attack and ensure the reliability of vehicles. First, the ML algorithm evaluates the vehicle message, and the result of this evaluation is used as feedback to combine with the Dempster-Shafer (DS) theory. Then the reputation score of each vehicle is calculated using DS theory with the combinational feedback of ML algorithm. Besides, this work proposes a revocation scheme to perform a reputation score update. The obtained score is synchronized with the CA to enhance confidence for misbehaviour detection. This work carries out extensive simulations of a realistic vehicular environment to create the dataset. A wide range of alert-specific features is collected including time, speed, position, distance, flow, and change/difference in these parameters. The authors evaluate their mechanism against false alert and position falsification attacks and perform a comparison with voting schemes. The proposed mechanism achieves an F1-score of 98\% for false alert and 96\% for positional attack detection. \par 
Tariq et. al \cite{CANTLIDS} present a TL-based intrusion detection scheme on CAN protocol. First, it trains the convolution LSTM-based model with a previously-known intrusion dataset. Later, one-shot TL is used to re-train the model for detection of new attacks where only one sample of new intrusion is enough to detect it. In the performance evaluation, the authors use the CAN dataset collected from two real vehicles to demonstrate 88\% and 95\% of accurate detection of new and known attack types, respectively. Another TL-based IDS is proposed in \cite{TLIDS}. In this work, the authors propose two TL-based model update schemes to detect new types of attacks in IoVs. It uses a tree-based TL algorithm along with two update schemes, a cloud-assisted and a local model update scheme. One of the key advantages of this work is using a small amount of data to achieve high detection accuracy. The model works with two hypotheses based on whether the cloud can timely provide labeled data or not. If the cloud provides data, the TL model update is performed with the given data. In case of no timely provision of data from the cloud, the local model is used. In the local update, vehicles obtain the pseudo-label of the unlabeled data and perform multiple updates of the TL model to respond to a new attack before the cloud completes the labeling of new attack data. This scheme achieves an accuracy of 92-96\% by evaluating two sets of data from the publicly available AWID (Aegean WiFi Intrusion) dataset. \par 
A recent work in \cite{MLDCMDS} presents a data-centric misbehaviour detection system for IoVs. The novelty of this work is about using plausibility checks along with traditional supervised ML algorithms to increase detection accuracy. The authors compare the performance of six supervised ML algorithms with two plausibility checks i.e. location plausibility and movement plausibility. The results show 5\% and 2\% of improvement in precision and recall, respectively, with the additional plausibility checks.

\subsection{Trust Computation}
\label{Sec:trust}

\begin{table*}[htbp]
	\centering
	\caption{Summary of Literature on Trust Computation using ML}
	\begin{tabular}{cP{1cm}P{3.5cm}P{3.5cm}P{2.5cm}P{2cm}}
		\toprule
		\textbf{Citation} & \textbf{Year} & \textbf{Feature Data} & \textbf{Algorithm} & \textbf{Type of Trust} & \textbf{Accuracy} \\
		\midrule
		\cite{ref69}  & 2016  & Basic safety messages & Logistic Regression & Data centric trust & NA \\

		\cite{ref61} & 2018  & Data Traffic & ML (5 schemes compared)  & Data centric trust & 100\% \\

		\cite{ref68} & 2018  & Routing information & RL & Data centric trust & NA \\
		
		\cite{nRef6} & 2018  & Routing information & DRL & Data centric trust & NA \\
		
		\cite{nRef7} & 2018  & Packet drop count, transfer delay and forward interval & SVM & Data centric trust & 98\% (Urban) and 91\% (Highway) \\
		\cite{ref66} & 2019  & Uniformaly distributed trust value & Q-Learning & Vehicle centric trust & 100\% \\

		\cite{ref67} & 2019  & Driver behaviour and received messages & DL & Hybrid trust & NA \\

		\cite{ref70} & 2019  & Similarity, Fimiliarity and PDR & Variants of KNN and SVM & Data centric trust & ~90\% \\
		
		\cite{ref81} & 2020  & Event, context and feedback from driving model & RL & Data centric trust & ~100\% \\
		\cite{nRef5} & 2020 & RSSI, PDR, and the distance between two vehicles & KNN & Hybrid trust & NA \\		
		\bottomrule
	\end{tabular}%
	\label{tab:trust}%
\end{table*}%

Trust is an important aspect of security to enhance the protection level of the system from attackers. In vehicular networks, trust computation is an additional step of security requirement used along with privacy protection, availability, and key management to ensure the highly-secure transmission of data \cite{ref84}. Trust computation schemes exist in the literature (in the context of entity-oriented and data-oriented) which make use of historic interaction of vehicles within a network to classify it as the trusted one \cite{ref85}. It is important in vehicular networks to ensure honesty among vehicles by performing trust computation as a basic security requirement. The summarized tabular view of solutions using ML for trust computation is shown in Table. \ref{tab:trust}. \par 

Ahmed et. al \cite{ref69} present an algorithm for identifying honest and dishonest nodes using logistic regression over the trust values of nodes. The trust values are computed from the messages. The more the correct messages are forwarded from the node, the higher is its trust value. Basic safety messages (BSM) are communicated between nodes that include speed, location, brake status, and other information related to the state of a vehicle. A sudden change in speed or brake may take place due to the fake information from a malicious node. These parameters are used to identify the correctness of the information and calculate trust values. Here, a logistic regression algorithm create trust values for all nodes and generates a list of honest and malicious nodes within a network. The authors use OMNET++ to simulate their proposed design. \par 

Shams et. al \cite{nRef7} present a trust establishment mechanism for vehicles, called Trust Aware SVM-Based IDS (TSIDS). It uses promiscuous mode to collect data, and SVM for classification of trusted vehicles. The packet drop count (PDC), packet transfer delay (PTD), and packet forward interval (PFI) are used as the feed of the classification module. Here, promiscuous mode enables capturing all data packets from all nodes in the reception range. This makes every node to monitor neighbouring nodes and detect misbehaviour in its surrounding. To save energy and processing power resources, the packet collection is initiated only during packet routing. The performance results show that the promiscuous mode-based SVM classifier detects trusted vehicles with an F-score of about 98\% for the urban scenario, and 91\% for the highway scenario.

In \cite{ref61}, the authors perform a comparison analysis between five different ML algorithms for the trust computation of data in machine-to-machine (M2M) communications. This work proposes to use the above ML-based solutions for vehicular applications (i.e. V2V communications). It uses MATLAB to simulate a scenario of connected vehicles/nodes for the evaluation of proposed models. The data transmitted between nodes is fed into the ML algorithm to evaluate the trustworthiness of data. This work observes that random forest achieves the best performance in terms of a receiver operation characteristic (ROC) and precision-recall curve (PRC) for evaluating the trust. \par 

Trust computation can also be used to perform secure routing in vehicular networks. Zhang et. al \cite{ref68} present the idea of using trust computation for reliable-routing using RL (i.e. with the use of rewards and actions) in SDN-based vehicular networks. In this work, each node attempts to find a trustworthy neighbour for data forwarding. First, V2V communication takes place to decide the trustworthiness of a vehicle, in terms of correctly-received packets depending on the total packets forwarded. Later, a deep CNN running over the controller calculates Q-values considering the network input state and an individual trust value as input to the model. Based on the calculated Q-values (a cumulative path trust value), a controller decides the forwarding path as an action to do next-hop routing of data. The convergence, PDR, and average network throughput are calculated to evaluate the effectiveness of an algorithm. In \cite{nRef6}, Zhang et. al extend \cite{ref68} and propose a trust-based dueling deep RL approach (T-DDRL) for routing data in SDVNs. In this model, a logically centralized controller deploys a dueling network architecture. The best policy to route data is determined using deep Q-learning. The controller acts as an agent that decides trusted immediate-path for routing, based on long-term rewards and Q-values. The authors use OPNET simulator to evaluate their model in terms of convergence performance, delay, and throughput. \par 

Recently, in \cite{nRef5}, Soleymani et. al perform trust computation using fuzzy logic along with an ML algorithm. It proposes a fuzzy logic-based trust computation mechanism to access integrity and accuracy in event messages and their sender. The parameters that are used to measure trust scores are plausibility, experience, and type of vehicle. The plausibility and experience are based on communication history and location. Here, the type of a vehicle is considered as low, medium, or high, based on its authentication level. This work considers obstacles to have a major impact while measuring trustworthiness. Therefore, it performs evaluations under line-of-sight (LOS)(i.e. a radio link between sender and receiver without an obstacle between them) and non line-of-sight (NLOS)(i.e. a radio link between sender and receiver with obstacles) environments. The differentiation of LOS and NLOS is done with the KNN algorithm by using RSSI, PDR, and the distance between two vehicles as input features. This work performs Monte-Carlo simulations and demonstrates that their work has better performance compared to other models under different patterns of attacks.

Trust computation is extremely important when it comes to driverless cars where human trust over the automated vehicles is still a question \cite{ref62}\cite{ref63}\cite{ref64}\cite{ref65}. The study in \cite{ref62} tries to learn the factors that may affect a driver's willingness to rely on an automated driving system. The authors evaluate the patterns of speed, lateral distance, and steering maneuver timings when an automated car is passing and/or overtaking a manual vehicle (scooter and bicycle are considered in this work). The driver responses (of the manual vehicle) are collected verbally and by providing a questionnaire. In terms of obtained responses, the proposed work highlights an appropriate range of the above factors in order to maintain a trust among vehicles which share the same road. However, in recent years, researchers are focusing on ML-based solutions to measure trust and implement actions to block dishonest nodes from entering the network. In \cite{ref66}, the authors employ a Q-learning algorithm to assess the trust for automated driving vehicles (ADVs) and report intruders, based on the measured confidence level. Two types of assessment methods are discussed in the proposed work which are called as direct and indirect ADV assessment models. In the direct model, trust is maintained based on V2V and V2R communications. In the indirect assessment model, a vehicle communicates with RSU and only RSU plays a role to ensure the reliability of a vehicle. The assessment values, collected for a vehicle, by other vehicles or different RSUs, are aggregated and used to confirm the misbehvaiour of a vehicle. A Q-learning model encourages vehicles to report intruders and in response vehicles receive rewards from the RSU to enhance their trust level. In performance evaluation, the authors prove the higher detection accuracy of their proposed model compared to the previous works. \par

Tangade et. al \cite{ref67} propose a DL-based algorithm to perform trust computation and driver classification in VANET. Two sequential DNNs are used in this work where the first network (5-layer) assigns reward points to the driver-based on the driving behaviour. The reward points also help to compute trust-value for different vehicles. However, the second sequential deep network (4-layer) performs computation to characterize the driver as fraudulent or non-fraudulent based on the received messages. The effective performance of their proposed algorithm in terms of fraudulent driver identification and low computational overhead is evaluated using NS-3 simulator. \par 

The authors in \cite{ref70} present a comparative analysis between different supervised ML algorithms to accurately detect trustworthy and untrustworthy nodes. An IoV environment is considered in this work. The trust level of different nodes is computed based on the direct and indirect observations of data transmitted by a node. However, the calculation of optimum weights and trust thresholds is considered in this work and measured by using ML techniques. A real IoT data set is used in this work to do performance evaluation. It extracts features such as similarity, familiarity, and PDR, and labels them to feed into 11 different variants of SVM and KNN supervised algorithms. The above ML techniques are evaluated in terms of precision, recall, F1-score, and accuracy, where most of the KNN variants are performing better than SVM. \par 

A data-oriented trust model is proposed by Guo et. al \cite{ref81} to help prevent driving decision-making entity from bogus information. In this model, a vehicle requests trust values from the trust evaluation model regarding a driving decision-making event such as path selection and speed regulation. The trust evaluation model connects to a data repository and learning engine module which deploys RL to optimize the trust evaluation decision by receiving feedback from the driving decision-making module of a vehicle. The data repository contains information such as onboard sensor readings received from different nodes at a different time and, previous trust decisions (feedback). This work demonstrates the use of feedback for optimizing the trust module process that yields a high precision rate in different scenarios. \par

\subsection{Privacy Protection}
\label{Sec:Privacy}

\begin{table*}[htbp]

	\centering

	\caption{Summary of Literature on Privacy Protection using ML}

	\begin{tabular}{cP{1cm}P{3.5cm}P{3.5cm}P{3.5cm}}

		\toprule

		\textbf{Citation} & \textbf{Year} & \textbf{Feature Data} & \textbf{Algorithm} & \textbf{Type of Privacy} \\

		\midrule

		\cite{ref79} & 2019  & Privacy level of last time slot, location coordinates and semantic location & RL & Semantic Trajectory (Location) \\

		\cite{ref80} & 2019  & Road images & FL & Data Privacy \\
		
		\cite{nRef1} & 2020  & NSL-KDD dataset & Collaborative Learning  & Training Data Privacy \\
		
		\cite{FLSecureData} & 2020 & MNIST dataset & FL, CNN and DRL & Data Privacy \\
		
		\cite{FLPP2} & 2020 & 20 News-groups dataset & FL and Gradient Boost Decision Tree & Data Privacy \\
		 
		\cite{PPFL} & 2021 & VeReMi dataset & FL & Data Privacy \\

		\bottomrule

	\end{tabular}%

	\label{tab:privacy}%

\end{table*}%

Privacy is a means of protecting the sensitive information of vehicles from attackers. In the context of vehicular networks, it is further classified into location privacy and user privacy. A wide range of privacy schemes are proposed in the literature which are categorized into mix-zones\cite{ref71}\cite{ref78}, obfuscation-based\cite{ref72}\cite{ref77}, silent-period\cite{ref73}, k-anonymity\cite{ref76}, dummy-based\cite{ref74} and hybrid models\cite{ref75}. Most of these techniques use a pseudonym (i.e. a unique identifier different from the real one) as an identity to get services from the network without exposing its real identification. The ability to access services without disclosing real identity is known as anonymity. In mix-zones and silent-period schemes, the location of a user is preserved by forcing it to change pseudonym using anonymous communication zones (i.e mix zones) where the density of vehicles is high and/or during specified silent periods, respectively. This makes it difficult for an attacker to map a pseudonym with a real identity. However, it's not important to always have high-density zones and a change of pseudonym at an inappropriate time may result in lack of privacy and performance degradation. In obfuscation-based mechanisms, the users obtain LBSs by using a path confusion algorithm in which the location of a nearby building, object, or intersecting user are used to obscure the real location. In this algorithm, the presence of suitable substitutes to create anonymity for the required LBS is not guaranteed. In k-anonymity and dummy-based mechanisms, the location of a user is mixed with k-nearby users or n-different dummy users before obtaining a service. These techniques involve maintaining a pool of locations which is difficult and processing-intensive. \par 
In recent years, ML-based schemes are receiving attention to overcome the drawbacks of the above-mentioned techniques. Protecting the privacy of users and locations from attackers while ensuring secure availability to the service-providing entity is quite challenging. New solutions based on ML can be developed for the new type of data such as images, generated from vehicles. Only a little work is done in the literature in this domain. The summarized tabular view of available ML-based solutions is presented in Table. \ref{tab:privacy}. \par 
Wang et al. \cite{ref79} propose an RL-based obfuscation scheme to enhance the privacy of the semantic trajectory of a vehicle. In this model, a vehicle communicates with RSU and provides location coordinates and semantic location to access LBSs. In order to protect the semantic trajectory, an obfuscated location is transmitted despite the real location coordinates. The selection of obfuscated location is dependent on the policy defined by an RL model. It observes the current state, privacy level of last time slot, real location coordinates, and semantic location to update its Q-function and give decisions for the most suitable obfuscation policy. The proposed work also assumes the adversary is smart and does not simply trust the location parameters received from a vehicle. It aims to minimize the vehicle's privacy gain and sends spams and scams by inferring the vehicle's semantic location and Markov model. This is why an RL-based mechanism selects the best obfuscation policy for a vehicle at any particular time while considering the adversary's behaviour. For the performance study, the privacy gain and quality of service loss are calculated to evaluate the functioning of the proposed model. \par 

A privacy-preserving technique using FL is proposed for IoV where data is generated in the form of images \cite{ref80}. This work focuses on the privacy-enhanced data collection at the edge of the networks. First, the data generated by a vehicle node is uploaded to an edge network. The FL pre-processes the data on an edge to estimate the road condition based on uploaded images. The function of the pre-processor is to perform image fingerprinting and to find correlation and similarity of images using semi-supervised learning. This also removes a large number of irrelevant images. Training results and some amount of pre-processed data are transmitted to the cloud (backbone network) for further processing and to share with other edge nodes. This scheme results in reduced delays and enhanced privacy of user data.\par 

Zhang et. al \cite{nRef1} propose to use collaborative learning in which vehicles share their experience with each other to assure better detection of malicious vehicles. Collaboration among vehicles also creates a privacy concern. This work presents a privacy-preserving ML-based collaborative IDS in vehicular networks. The idea is to detect intrusion while maintaining the privacy of the training dataset. It uses alternating direction method of multipliers (ADMM) to train a classifier for detecting misbehaviour from vehicles. It uses a supervised learning algorithm and each collaborative learning is modeled as an optimization problem to perform distributed ADMM-based empirical risk minimization (ERM). In addition, a dual variable perturbation (DVP) is applied to preserve the privacy of the training dataset. The authors evaluate their design in terms of empirical risk and empirical loss to quantify security for different network topologies. \par
Lu et. al \cite{FLSecureData} address provider's privacy concern and present a blockchain-integrated FL framework for secure data sharing. It proposes to use two directed acyclic graphs (DAGs), one main permissioned blockchain at RSU (i.e. PermiDAG), and a local DAG at vehicles for secure data sharing. Moreover, a DRL-adopted asynchronous FL framework is used to perform efficient node selection. The learning models will also participate in enhancing the reliability of data shared between an RSU and vehicles by executing a two-stage verification. The performance results show faster convergence and higher accuracy of the proposed learning models. \par 
Lu et. al \cite{FLPP2} propose a two-phase federated learning-based data privacy protection scheme for vehicular cyber-physical systems (VCPS). It consists of data transformation and collaborative data leakage detection. First, it allows vehicles to locally-train models and ensure privacy of their data. This work also addresses the limited resource problem, and to save computing resources, it performs caching of a trained model in each phase for repeated use. Later, a federated scheme-based mechanism is used to address the vulnerability problems at a centralized node. In this work, the authors use the gradient-descent algorithm and Laplace mechanism to formulate a learning problem and distort the model for guaranteeing differential privacy. To evaluate the model, this work uses a real-world 20 News-group dataset and demonstrates high security, high efficiency, near-real-time performance and good accuracy of the proposed scheme.\par 
A recent study on privacy protection using federated learning is presented in \cite{PPFL}. The idea of this work is to learn a misbehaviour detection model while maintaining user data privacy. Vehicles use BSMs to exchange data such as speed and location, which help ML algorithms to make routing, guidance, and safety decisions. In this work, the authors study data falsification attacks that compromise the privacy of personal information transmitted using BSMs between vehicles. As a solution, it proposes federated learning where personal information of a vehicle resides locally on the vehicle and performs ML training without sending data to the central node. The vehicles only send their updated and trained local model to the central node for learning of an aggregated smarter model. The centrally trained model not only detects but also identifies the position of an attack. In the performance study, the authors demonstrate the effectiveness of federated learning compared to the centrally BSM-trained model.

\section{Limitations and Challenges and in Using ML-based Security Solutions}
\label{Sec:MLlimitations}
ML has shown significant achievements and is becoming a workhorse for many security applications in vehicular networks. At the same time, it has a number of limitations and constraints. This section highlights the limitations in ML-based solutions for vehicular networks and brings out the research challenges that need to be addressed.
	
\subsection{Adversarial Machine Learning}
Adversarial ML is an important limitation of using ML-based solutions to different security problems in vehicular networks. In adversarial ML, the adversaries may use multiple ways to supply deceptive inputs to the ML model, attempt to fool it, and compromise the results \cite{AML}. With the advent of vehicle automation which involves the use of ML and multi-agent systems to assist a vehicle's operation, an adversarial attack over ML may result in catastrophes and great danger to human lives. Therefore, dealing with adversarial ML is critically important to ensure the robustness of algorithms proposed in the literature to protect vehicular nodes and related information, as discussed in Section \ref{Sec:LearningSolutions}. In this section, we briefly discuss the threats to different types of ML techniques used to secure vehicular networks.   

\subsubsection{Threats to Supervised Learning}
The conventional machine learning algorithms where classification takes place using static features and predefined labels are vulnerable to deliberate attacks. Evasion attack is one of the most common types of deliberate attacks over supervised ML algorithms. In an evasion attack, an adversary aims to manipulate test samples that are undetectable by ML classifiers\cite{EvasionAttack}. There are several vehicular applications where ML-based supervised classification is used to differentiate malicious and non-malicious nodes, as discussed in Section \ref{Sec:LearningSolutions}. In an evasion attack, a vehicle (adversary) can easily add malicious test samples without participating or changing training data and induce an ML algorithm to output incorrect results. In \cite{SupervisedMLAttack_Vehicle}, authors perform an experimental study over a vehicular network to fool its supervised model and demonstrate how ML-generated attacks over ML are undetectable by existing ML classifiers in vehicular applications. \par 
The feature poisoning in ML algorithms is another threat to supervised algorithms. Different feature selection/extraction techniques are exploited in the literature for diverse vehicular applications, including attack prevention and intrusion detection\cite{Fattack1,Fattack2}. The selection of the right features not only reduces the computational cost but also improves the learning capabilities of the algorithm. In the case of high-dimension datasets, it is not effective to perform manual feature selection. Therefore, automated feature selection methods are useful. However, such automated techniques are unfavorable when training features are poisoned by smart attackers \cite{featureSelectionAttack}. In \cite{featureSelectionAttack}, the authors perform a study over some popular feature selection methods to show how easily these methods can be compromised under positioning attack over training samples. \par 
Therefore, reliable deployment of supervised algorithms in security-sensitive vehicular networks is important. A secure ML-classifier with a robust feature selection method can be designed to enhance regulatory terms for the legitimate users and to differentiate between the legitimate and illegitimate training entries.

\subsubsection{Threats to Unsupervised Learning}
The problem of attacks over unsupervised learning is not explored in the context of vehicular networks. Nonetheless, unsupervised algorithms are also susceptible to adversarial attacks \cite{UnsupervisedAttacks}. Like supervised learning, in unsupervised learning, evasion attack is also effective where the addition of malicious samples during the testing phase could lead to learning vulnerabilities evading detection of an attacker, or attacking clustering method by reducing the distance between real and adversarial samples will mislead the model, and classify it under the class of legitimate users.  However, such attacks, their impact, and their prevention in the context of vehicular applications are not explored much in the literature. This opens up a direction for researchers to shed light on the area of adversarial attacks over unsupervised mechanisms in vehicular network security.

\subsubsection{Threats to Reinforcement Learning} 
Reinforcement learning and its DL-assisted variants aid autonomous vehicles to make self-directed decisions in driving task\cite{AD1,AD2,AD3}. Despite the development, RL may suffer from various issues in autonomous systems such as wrongly-classified objects in perception systems, theft of vehicles due to incorrect recognition of driver monitoring patterns by a neural network, compromised functional safety due to erroneous collection of vehicle data, and failure in the detection of an attack. In RL, where no prior knowledge exists, and actions are derived based on long-term rewards, an agent may keep on increasing rewards for a fooled environment in the event of an attack\cite{RLFool}. In addition, recovery time in RL algorithms is generally high. Therefore, several researchers have raised concerns regarding reliable RL for security-critical applications \cite{AdvRL1,AdvRL3,AdvRL4}.\par
The authors in \cite{RLAttack2} explore the adversarial attacks on DRL. Wang et. al \cite{RLAttack2} investigate the impact of adversaries on a well-trained DRL-based energy management electric vehicle. The attacks are generated using the fast-gradient-sign method (FGSM) where different assumptions are generated against DRL, like too much usage of fuel or out of battery, to confuse the energy management system and distract the performance of the electric vehicle. This work shows significant performance degradation of targeted DRL agent against two types of adversarial random noises. Therefore, a robust method that monitors inputs before processing is needed. In another work, Yue Wang et. al \cite{RLAttack1} explore backdoor trojanning attacks on DRL-based congestion control system of autonomous vehicles. The authors investigate a set of triggers to enhance the stealthiness of the attack before injecting it into the DRL training set and ensures the similarity of triggers to the benign data. Later, the backdoor injection takes place by retraining the model with a mixture of legitimate and malicious entries. Here, the trigger set consists of vehicle position and speed, and malicious actions are acceleration and deceleration control. The performance evaluation of the attack model is done using three different complex traffic scenarios. The backdoor injection tricks the system toward an insurance attack, where an autonomous vehicle crashes into the vehicle in front of it.\par
Not limited to this, the addition of noise to RL inputs for distracted graphical perception, injecting faulty data to sensor readings to misguide the agent, and/or physical attack where an adversary may alter the physical environment around vehicles, is another easy-to-deploy adversarial attack on a DRL model. This opens up several opportunities to develop efficient, robust, interpretable, fair, and defensive RL for autonomous vehicles and other vehicular applications.

\subsubsection{Threats to Deep Learning}
Recently, there are several works in the literature which use DL for different applications in vehicular networks including, security\cite{DNNVehicle}\cite{DNNvehicle2}, autonomous vehicle control system\cite{DNNvehicle3}, traffic light control\cite{DNNvehicle4}, and so on. However, DL agorithms can easily be fooled using several techniques \cite{AttacksOnDL,ML_Attack,ML_Attack2,ML_Attack3,ML_Attack4}. A survey is presented in \cite{AttacksOnDL} which highlights the problem of an adversarial attack on DL as a serious issue for networks.\par
In driverless/autonomous cars, it is very crucial to understand the environment to make correct decisions. An autonomous vehicle perceives the nearby objects and prepares its path trajectory based on sensors such as cameras. The use of CNN and DNN is highlighted in most of the works to deal with an image dataset \cite{CNNImage,CNNImage2,CNNImage3,CNNImage4}. While DNN and CNN deal with image data efficiently, it can easily be fooled by an attacker through fake scenes which results in noisy images and false predictions. A NVIDIA PilotNet architecture in autonomous vehicles estimates steering angles based on the perception received from camera sensors. In \cite{Pilotnet}, authors demonstrate learning and mapping using DNN in NVIDIA PilotNet architecture. Although DNNs are useful, they are relatively easy to be fooled with high confidence, which may result in serious degradation of network performance by predicting wrong outputs in autonomous vehicular applications. In \cite{DNNCNNFooling}, the authors produce noisy images (fooling images) using gradient ascent which are unrecognizable to humans, but a trained DNN model recognizes it as belonging to one of its classes with high confidence. Such a type of fooling adversarial attacks may result in catastrophes and great danger to human lives in the case of a fully autonomous vehicle.\par
Therefore, dealing with such image-fooling attacks becomes critically important to ensure the robustness of deep algorithms, and mitigate the impacts of fooling images. Although researchers have started addressing issues with image data, approaches which do not use images as input, are yet to be explored. This opens up a new research direction for the researchers to study all types of data inputs and propose robust and safer deep architectures for self-driving cars and other vehicular applications, under adversarial conditions.

\subsubsection{Threats to Federated Learning}
Federated learning is recently being explored for privacy protection and many other applications in vehicular networks \cite{FLPS1,FLPS2}. It provides highly-centralized services for vehicles and follows the pattern where global learning (at a central node) and local learning (at vehicles) exchange sensitive data over the air (wireless channel) to offer fast training, low processing and, low communication overhead \cite{FDL}. To protect model confidentiality, the global node receives no information about how the data is generated from local nodes. This makes federated learning less-prone to model poisoning attacks. However, it is possible to launch a backdoor attack which is explored in the literature against FL, where an adversary performs model replacement at the local level to interrupt the global performance \cite{AttackOverFL1,AttackOverFL2}. Therefore, ensuring the reliability of participating nodes is important. The vehicular networks are mobile where network configuration changes quickly. As a result, participating nodes also change until the FL model convergences, and the inclusion of unreliable nodes in-between may jeopardize the performance of an FL model.\par 
Another possible attack on FL is data poisoning attack. An adversary can launch a data poisoning attack over the data transmitted wirelessly to alter the prediction/classification behaviour of the FL algorithm. The sensitive information transmitted between local and global models includes weights, bias, and gradient values. This requires local nodes to implement additional privacy algorithms. However, attacks on FL in the area of vehicular applications and its impact on user security is yet an area to explore in the future. Moreover, the protection of FL leads to additional computational overhead whereas vehicular nodes are resource-constrained entities. The solutions in this context require intelligent algorithms which are smart enough to understand the trade-off between available resources and privacy needed to prevent the attack on FL model or data.

\subsubsection{Threats to Transfer Learning}
Transfer learning is considered as a powerful approach due to its ability to quickly build a new ML model from the existing pre-trained model. In the context of vehicular security, the use of transfer learning is observed in intrusion detection and attack detection applications. Recently, researchers have done some experimental studies to validate the vulnerability of transfer learning toward misclassification attacks, weight poisoning attacks and backdoor attacks \cite{AttackonTL1,AttackonTL2,AttackonTL3}. Here, a misclassification attack tries to manipulate the features of certain layer outputs, whereas, in weight poisoning, the adversary injects vulnerability into pre-trained weights. In the backdoor, the attacker aims to craft an adversarial model from a pre-trained model to manipulate the end-to-end classification system. However, the impact of such attacks and the level of catastrophe it may create are not studied for vehicular networks and the users. Therefore, this area is still an open issue to explore further.

\subsection{Energy Constraint in ML-based Solutions}
Energy is another important constraint that needs to be considered while proposing ML-based security solutions. Fast training and accurate detection are key concerns in ML for an efficient design. However, it can only be achieved with large data and heavy machinery which means high energy consumption and more system resources. Therefore, ML algorithms are regarded as energy-consuming resource-intensive solutions. Recently, the advent of autonomous vehicles has driven research on the problem of the security of in-vehicle network. An in-vehicle network means battery-powered sensors and an embedded transceiver unit. Performing ML training locally on the vehicle (as done in \cite{FLPP2,PPFL}, Section \ref{Sec:Privacy}) can be energy expensive. In addition, the heterogeneity of devices in vehicular networks, such as pedestrians with a smartphone requires optimization of an algorithm according to the type of a node. This is because, unlike a vehicle, energy and resource efficiency is critical for the pedestrian user. Hence, an energy constraint is important to study while providing ML-based security solutions in vehicular networks. As discussed in Section \ref{Sec:LearningSolutions}, recently, there are a few works done in the domain of energy-efficient secure ML designs where RL and TL are used in all of the works for attack detection in vehicular networks \cite{RLDOS,ref55,ref38,RLAttack2}. The constraint of energy efficiency is not much explored by other ML-based solutions used in the application of vehicular security. Withal, data-driven and energy-efficient ML techniques are a challenge that need to be considered in vehicular security.\par 
Another potential solution is resource offloading which is explored by researchers lately for enhancing the energy performance of future ML algorithms. Compounded by high mobility and a high number of users, offloading is not an easy-to-solve problem in vehicular networks, and offloading ML-based security algorithm and their performance impacts is yet an open research issue to consider. \par 
Not limited to this, as discussed in Section \ref{Sec:LearningSolutions}, recently authors in \cite{FLSecureData} present the idea of using blockchain-integrated ML solution for privacy protection in vehicular application. With the advantage of authentication and data protection, blockchains are also known for the drawbacks of harder scalability, high energy dependence, and high resource consumption\cite{Blockchain2}. Therefore, energy-efficient and memory-efficient blockchain-integrated ML protocols for storing and processing blockchain consensus and blockchain-centric vehicular architectures are some important areas that open up new opportunities for further research.

\subsection{Latency Limitation in ML-based Solutions}
The latency has become a key driver for vehicular applications. ML-based solutions exhibit iterative execution property which results in longer time to generate outcomes. This makes latency limitation in ML for vehicular security solution, a potential open research issue, and there is a need to optimize the execution and response time of ML-based solutions. In addition, as discussed in Section \ref{Sec:attackDetection} and \ref{Sec:trust}, latency/delay is used as a metric in the number of security solutions to evaluate vehicle behaviour and performance of the ML-approaches.\par 

Application-differentiated ML approaches, that can characterize the type and requirements of applications, and quantify the parameters needed to deploy low-latency ML solutions, are desirable. As an example, latency requirements for ML-based autonomous driving security are different and more critical than the security of entertainment services in vehicles. Delays in detecting attacks over driving data can result in catastrophes or accidents in vehicles.\par 

A potential approach for a low-latency ML-based vehicular security is integration of ML and edge computing. Edge computing is becoming a key enabler for improved performance of the vehicular networks where edge computing seeks to reduce latency by migrating cloud services to an intermediate node that is closer to the vehicles. As discussed in Section \ref{Sec:LearningSolutions}, a recent work in the literature uses edge resources for fast ML-based security solutions\cite{FLPP2,ref55}. Considering this, deployment of ML techniques at edge nodes is an interesting area to explore which will help to make real-time responses by learning data locally. ML at the edge can solve security concerns by reducing reliance on the cloud network which requires personal vehicular information to be transmitted and processed at another end of the network. With edge, such crucial and private data can be processed locally in a real-time and faster manner which is not possible with the existing traditional network architectures. However, the drawback of using ML at the edge means working with data solely available on a single node, whereas centralized cloud means global knowledge. Moreover, there are techniques in ML (i.e. transfer learning) where models trained locally at the edge can be shared/transferred among nodes to gain more knowledge and with less data (model only) to be transferred. The more the models combine, the more the knowledge gained, and higher the quality of decision. At the same time, it could result in increased time for overhead. Thus, there is a tradeoff that needs to be carefully considered. \par 
In addition, edge networks are resource-constrained and with the limited resources available at the edge, the storage and computation resources required by ML algorithms may not be enough. Therefore, an intelligent technique to deploy a low-latency ML algorithm at the edge of a geographically diverse vehicular network is a potential area to explore further.

\subsection{Computation Cost in ML-based Solutions}
Vehicular networks are resource-constrained and mobile in nature. The use of DL-based security solutions requires more computational resources and a large amount of input data to achieve better optimization. The computation-intensive ML is another limitation in vehicular security. Some work is done in the literature which focus on computation-efficient ML-based solutions for vehicular security. As an example, recently, authors in \cite{CANTLIDS} (as discussed in Section \ref{Sec:MDS}) propose a light-weight and low-computation DL model for intrusion detection in vehicular networks. This work uses a one-shot TL over a pre-trained supervised LSTM  model for the detection of new attacks where only one sample of new intrusion is enough to detect it. Another recent work in \cite{TLIDS} proposes tree-based TL architecture for IDS which is a lightweight and low-computation model that achieves an accuracy of up to 96\%. This shows transfer learning has the potential of achieving computation efficiency in ML-based security solutions. However, with only a little work done, further studies and research can be carried out on the use of TL and its performance impacts for the application of vehicular security. \par 
Another potential research issue on computation-efficient ML is hardware-for-ML. It is a new trend recently seen in the literature \cite{MLHardware}\cite{MLHardware2}. An Eyeriss accelerator with the spatial architecture of 168 processing elements for deep convolutional architectures (AlexNet and VGG-16) is proposed by MIT \cite{MLHardwareMIT}. It optimizes computation efficiency by compressing and reusing local data using the dynamic random-access memory (DRAM) and accumulation unit. Many other research organizations such as, Google \cite{MLHardwareGoogle}, Stanford \cite{MLHardwareStanford} and IBM \cite{MLHardwareIBM} are also introducing hardware-based ML units. They support domain-specific hardware to provide robust, energy-efficient, and computation-efficient ML solutions. A potential future direction of research is to design secure vehicular network-friendly ML hardware. A low-processing ML-based hardware unit will help to provide a well-grounded computation-efficient performance over the vehicular access network.

\section{Observations and Lessons Learned}
\label{Sec:Observations}
In this section, we present the observations and lessons learned from the works presented in this survey on ML-based solutions for vehicular network security.\par 
In Section II, we highlighted the emergence of modern technologies such as 5G, SDN, edge computing, and cloud computing which has enabled different applications leading to the creation of new variants of vehicular networks. The key advantages of new variants are summarized in Table \ref{tab:VariantADV}. In the context of machine learning and security, we observe VANET and IoV are the most popular and widely used network architectures, whereas 5G, edge-enabled and cloud networks are recently gaining attention in vehicular applications.\par 
	\begin{table}[htbp]
	  \centering
	  \caption{Vehicular Network Variants and Advantages}
	    \begin{tabular}{| l | L{6cm} |}
	    \hline 
	    \textbf{Variant} & \textbf{Advantages} \\
	    \hline
	    IoV & Intelligent; allow large scale deployment; and integrate an environmental understanding of surrounding things such as human (driver) actions and activities \\
	    \hline
	    5GVN  & Provide high speed, low latency, efficient solutions for congestion control, fair resource sharing, reliability, high-throughput, high-connectivity and support diverse security applications \\
	    \hline
	    SDVN  & Handle the dynamic nature of the vehicles and supports better QoS, routing reliability, and security of the nodes. \\
	    \hline
	    EEVN  & Ideal for low-delay applications, performing scheduling and improved QoS \\
	    \hline
	    VCC   & Access to virtual services to road users and minimize onboard storage and computation \\
	    \hline
	    \end{tabular}%
	  \label{tab:VariantADV}%
	\end{table}%
It is observed that data transfer in vehicular networks involves a wide range of communication types. These types are susceptible to various types of vulnerabilities and security problems. We note that most of the ML-based literature focus on security in V2V, V2I, and V2R communication. Recently, the advent of autonomous vehicles has made the security of in-vehicle networks an important concern and researchers have started to explore this area. Several ML-based works have been carried out on the problems of availability, integrity, authentication, and trust computation. In the context of privacy, the use of ML is at an infant stage and is receiving attention recently. The use of ML for achieving confidentiality is not much studied in the literature, and it appears that most of the work use key-based approaches which seem enough to achieve confidentiality within in a system\cite{KeyManage1,KeyManage2,KeyManage3,KeyManage4}.\par 
In Section III, we reviewed the functioning of several machine learning techniques that serve as an analytical framework in various vehicular applications. Supervised and Unsupervised architectures work with various types of pre-collected datasets. However, reinforcement learning uses different policies and has the ability to collect and learn data in parallel where an agent generates data and maximizes the rewards by interacting with an environment. The use of RL is efficient in dynamic and mobile environments (such as vehicular networks) where the set of actions are infinite. We observed that advances in the functionalities have evolved into several other learning types such as DL, TL, and FL. \par 
DL does not require feature engineering and processes raw data in its original form. It is well-suited for the non-linear data patterns and it works in a supervised, partially-supervised, or unsupervised manner. For the cases of multi-dimensional data, CNN works good, but in applications where understanding temporal information is required, RNN is a better alternative to CNN. RNN is specially designed to process sequence data and/or time-series data. FL and TL are new types of learning strategies, work in conjunction with the existing ML models and recently showing good potential in the field of vehicular security. FL is a distributed and lightweight learning method wherein different networks at the lower level are trained based on the obtained weights from the global (centred) model to support ensemble learning. FL helps to reduce communication and processing overhead. However, TL is efficient in reducing the training time for a network. As the name suggests, it transfers important parameters among networks to fine-tune its working mechanism and the network does not go through the whole process of training again. TL also finds useful applications in mobile networks such as vehicular networks, where state changes occur frequently.\par 
In Section IV, we have identified the use of ML-based solutions in five different security aspects including, driver fingerprinting, attack detection, misbehaviour or intrusion detection, trust computation, and privacy protection. Several ML architectures are explored in these areas where we observe supervised learning, reinforcement learning, and deep learning are popular learning algorithms in vehicular security applications, as shown in Table \ref{tab:MLCompObservation} and \ref{tab:MLCompObservation2}. In the context of driver fingerprinting, it is of utmost importance to preserve the privacy of driver data and prevent adversaries from tracking the vehicle by linking it to the driver's identification. The older techniques use the idea of hiding location and accessing resources/services without disclosing the true identity (i.e, using pseudonymity) of a driver to ensure privacy. With the advent of autonomous vehicles, true profiling of the driver is important. To enable this, ML techniques are used by researchers to authenticate the true identification and fingerprinting of a user/driver to make vehicles resilient to theft without hiding the true identity of drivers. We also note that supervised learning is the most-widely used ML technique for driver fingerprinting, as shown in Table \ref{tab:MLCompObservation}. Some works in the literature, use unsupervised ML to cluster driver behaviour before performing classification and achieve better accuracies with the given models. It is observed that the data from smartphone sensors, vehicle sensors, and onboard electronic control units (i.e. using the OBD-II protocol) are the most commonly used data types for driver profiling using ML. \par 
In the context of attack detection, the designs of ML and deep architectures are attractive to detect a variety of attacks through experience and shared information. We observe that all three types of classes, including supervised, unsupervised, and reinforcement learning are beneficial in the detection and prevention of attacks. It is also noticed that the attacks over the availability and authenticity of vehicular networks are the most common attacks detected with ML-based solutions. In addition, the use of deep architectures is explored in the literature to maintain the integrity of platoons. In terms of data, we find that the wireless channel measurement data and routing traces are widely used in different types of attack detections. The advent of autonomous vehicles where sensors make some of the driving-related decisions requires a highly robust control system. The researchers consider reinforcement learning as a key player in this area to prevent different types of attacks on the sensors. In such cases, sensor data is used as a feature to detect and/or prevent the attack. \par 
Not limited to the known types of attacks, several ML-based studies have been carried out in the literature for the detection of unknown types of attacks i.e. IDS/MDS. We note that the supervised learning, deep learning, and transfer learning are the commonly used techniques in IDS. To collect feature set, a diverse set of data modalities where vehicular application-specific datasets like NGSIM, VeReMi, and CAN bus are used in recent studies for IDS and privacy protection. We also observe from the literature that datasets such as KDD-CUP, NSL-KDD, AWID, and CIC IDS, which are the application, protocol, and low-level network entities datasets, and not relevant but used for the performance study of intrusion detection and privacy protection in vehicular networks. However, it is highly desirable to use vehicular network specific datasets to provide more useful insights.\par 
The problem of privacy protection and trust computation using ML is relatively new and recently explored in the literature, and yet an open issue to explore further. Federated ML along with RL and DL is a recent and widely explored technique for privacy in vehicular networks. Another interesting observation is the use of blockchain-integrated ML solutions in vehicular privacy. In general, blockchain-based systems are playing an important role in authentication, privacy preservation, trust management, data management, and resource sharing applications \cite{Blockchain2}. However, blockchain-integrated learning frameworks open new directions for researchers, especially in 5G-based vehicular networks, where an increasing number of users may result in a larger block size, leading to network congestion affecting the power usage.\par 
The trust computation in vehicular networks is another basic security requirement which is used along with privacy protection, availability, and key management to ensure highly-secure transmission of data. In the literature, researchers make use of the historic interaction of vehicles within a network to classify it as the trusted one. The use of supervised ML, DL, and RL is explored widely in the literature with routing and wireless channel measurement data to ensure honesty among vehicles by performing trust computation.\par 
	\begin{table}[htbp]
		\centering
		\caption{ML techniques applied to vehicular security (Part 01)}
		\begin{tabular}{P{0.8cm}P{0.8cm}P{2.5cm}P{3cm}}
			\toprule
			\textbf{Citation} & \textbf{Year} & \textbf{Learning Technique} & \textbf{Security Application} \\
			\midrule
			\cite{ref51} & 2011  & \multirow{35}[1]{*}{Supervised learning} & MDS \\
			\cite{ref10} & 2013  &  & Driver Fingerprinting \\
			\cite{ref45} & 2015  &       & IDS \\
			\cite{ref47} & 2015  &       & IDS \\
			\cite{ref58} & 2015  &       & IDS \\
			\cite{ref9} & 2015  &       & Driver Fingerprinting \\ 
			\cite{ref69} & 2016  &       & Trust Computation \\
			\cite{ref59} & 2016  &       & IDS \\
			\cite{ref26} & 2016  &       & Grey Hole and Black Hole Detection\\
			\cite{ref11} & 2016  &       & Driver Fingerprinting \\
			\cite{ref12} & 2016  &       & Driver Fingerprinting \\
			\cite{ref48} & 2017  &       & IDS \\
			\cite{ref27} & 2017  &       & Sybil Attack Detection\\
			\cite{ref30} & 2017  &       & Sybil Attack Detection\\
			\cite{ref14} & 2017  &       & Driver Fingerprinting \\
			\cite{ref15} & 2017  &       & Driver Fingerprinting \\
			\cite{ref16} & 2017  &       & Driver Fingerprinting \\
			\cite{ref25} & 2018  &       & DDoS Detection \\
			\cite{ref43} & 2018  &       & MDS \\
			\cite{ref61} & 2018  &       & Trust Computation \\
			\cite{nRef7} & 2018  &       & Trust Computation \\
			\cite{ref19} & 2018  &       & Driver Fingerprinting \\
			\cite{ref20} & 2018  &       & Driver Fingerprinting \\
			\cite{ref46} & 2018  &       & DDoS Detection\\
			\cite{ref70} & 2019  &       & Trust Computation \\
			\cite{ref40} & 2019  &       & Jamming Attack Detection\\
			\cite{ref31} & 2019  &       & MDS \\
			\cite{ref54} & 2019  &       & Spoofing Attack Detection\\
			\cite{SupMDS} &  2019 & & MDS \\
			\cite{nRef3} & 2020  &       & Driver Fingerprinting \\
			\cite{nRef8} & 2020  &       & MDS \\
			\cite{nRef1} & 2020  &       & Privacy Protection \\
			\cite{nRef4} & 2020  &       & MDS \\
			\cite{nRef5} & 2020  &       & Trust Computation \\
			\cite{MLDCMDS} & 2021 & & MDS \\
			\midrule
			\cite{ref8} & 2012  & \multirow{8}[0]{*}{Unsupervised Learning} & Driver Fingerprinting \\
			\cite{ref10} & 2013  &  & Driver Fingerprinting \\
			\cite{ref39} & 2016  &  & MDS \\
			\cite{ref16} & 2017  &       & Driver Fingerprinting \\
			\cite{ref17} & 2017  &       & Driver Fingerprinting \\
			\cite{ref50} & 2018  &       & Jamming Attack Detection\\
			\cite{ref24} & 2018  &       & Manipulation Attack Detection\\
			\cite{USAttack} & 2020 &   & Attack Detection \\
			\midrule
			\cite{nRef6} & 2018  & \multirow{12}[0]{*}{Reinforcement Learning} & Trust Computation \\
			\cite{ref68} & 2018  &       & Trust Computation \\
			\cite{ref57} & 2018  &       & Spoofing Attack Detection\\
			\cite{ref38} & 2018  &       & Jamming Attack Detection\\
			\cite{RLCPAttack} & 2018 & 	& Cyber-Physical Attack Detection \\
			\cite{ref79} & 2019  &       & Privacy Protection \\
			\cite{ref36} & 2019  &       & Jamming Attack Detection\\
			\cite{ref66} & 2019  &       & Trust Computation \\
			\cite{RLDOS} & 2019 & 	   &  DDoS Attack Detection \\
			\cite{ref55} & 2020  &      & Spoofing Attack Detection \\
			\cite{ref81} & 2020  &       & Trust Computation \\
			\cite{FLSecureData} & 2020 & & Privacy Protection \\
			\bottomrule
		\end{tabular}%
		\label{tab:MLCompObservation}%
	\end{table}%
	
	\begin{table}[htbp]
		\centering
		\caption{ML techniques applied to vehicular security (Part 02)}
		\begin{tabular}{P{0.8cm}P{0.8cm}P{2.5cm}P{3cm}}
			\toprule
			\textbf{Citation} & \textbf{Year} & \textbf{Learning Technique} & \textbf{Security Application} \\
			\midrule
			\cite{ref32} & 2015  & \multirow{15}[0]{*}{Deep Learning} & Black hole \\
			\cite{ref12} & 2016  &       & Driver Fingerprinting \\
			\cite{ref13} & 2016  &       & Driver Fingerprinting \\
			\cite{ref34} & 2017  &       & MDS \\
			\cite{ref18} & 2017  &       & Driver Fingerprinting \\
			\cite{DLIDS} & 2018  &       & IDS \\
			\cite{ref21} & 2018  &       & Driver Fingerprinting \\
			\cite{RLCPAttack} & 2018 & 	& Cyber-Physical Attack Detection \\
			\cite{ref22} & 2019  &       & Platoon Attack Detection\\
			\cite{ref33} & 2019  &       & Sybil Attack Detection\\
			\cite{ref53} & 2019  &       & Crossfire Attack Detection\\
			\cite{ref67} & 2019  &       & Trust Computation \\
			\cite{Ref5} & 2019  &       & Driver Fingerprinting \\
			\cite{nRef2} & 2020  &       & IDS \\
			\cite{DLCPAttack} & 2020 & 	& Data Manipulation Attack Detection \\
			\midrule
			\cite{ref80} & 2019  & \multirow{4}[1]{*}{Federated Learning} & Privacy Protection \\
			\cite{FLSecureData} & 2020 & & Privacy Protection \\
			\cite{FLPP2} & 2020 & & Privacy Protection \\
			\cite{PPFL} & 2021 & & Privacy Protection \\
			\midrule
			\cite{CANTLIDS} & 2020  &  \multirow{3}[1]{*}{Transfer Learning}     & IDS \\
			\cite{ref55} & 2020  &       & Spoofing Attack Detection\\
			\cite{TLIDS} & 2021  &  & IDS \\
			\bottomrule
		\end{tabular}%
		\label{tab:MLCompObservation2}%
	\end{table}%
In Section V, we have observed that despite the impressive achievements of ML in vehicular security applications, it has a number of limitations and challenges. This section also discusses the possibilities of further research in ML-based vehicular security. One important limitation is adversarial ML. In several ways, the vehicles (adversaries) may supply deceptive inputs to the ML model in an attempt to fool it, and compromise the results. The conventional supervised algorithms where classification takes place using static features and predefined labels are vulnerable to feature poisoning and deliberate attacks where an adversary aims to manipulate training features and test samples, respectively, and induce an ML algorithm to output incorrect results. The area of attacks over unsupervised learning is not explored much in the context of vehicular networks. Nonetheless, we observe that unsupervised algorithms are also susceptible to adversarial attacks. Therefore, reliable deployment of supervised and unsupervised algorithms with a robust feature selection method in security-sensitive vehicular networks is important.\par 

We highlight that RL, an attractive technique for autonomous vehicles, may suffer from various issues such as wrongly classified objects in perception systems, theft of vehicles due to incorrect recognition of driver monitoring patterns by the neural network, compromised functional safety due to erroneous collection of vehicle data, failure in the detection of an attack and many more. In the event of an attack over RL, an agent may keep on increasing rewards for illegitimate actions, resulting in a fooled environment. Therefore, the recovery time to go back to normal state becomes high. Some researchers have raised concern over the reliability of RL for vehicular security applications. Therefore, there is a need to put future efforts into designing defensive RL for vehicular applications.\par 

For DL-related security, many vehicular applications deal with image datasets and such models can easily be fooled by an attacker through fake scenes which results in noisy images and false predictions. The image fooling adversarial attacks may result in catastrophes and great danger to human lives in the case of autonomous vehicles. We also observe that, while FL is less effective to model poisoning attacks, backdoor attacks and data poisoning attacks are common. The backdoor attack along with misclassification attack and weight poisoning attack is also studied to validate the vulnerability of transfer learning. The attacks on FL and TL in the area of vehicular applications and their impact on user security are less explored. The extent of catastrophe caused by the above types of attacks over different ML types is not studied much in the context of vehicular networks.\par

Not limited to adversaries, there are some other key constraints to study while developing ML-based security solutions. The limitations discussed here are latency, energy consumption, resource usage, and computation efficiency. The above limitations are related to each other in one way or another, and choosing a right tradeoff among them is also an important problem to study. The possible solutions we discussed are the use of data-driven algorithms, well-quantification of application sensitivity, edge-ML, resource offloading, blockchain integrated ML, new ML techniques like TL, and hardware-for-ML. All these solutions are less explored and have good potential for further research in vehicular security.\par  

\section{Conclusion}
\label{Sec:Conclusion}
ML techniques offer huge benefits to enable secure communication in vehicular networks. High mobility of vehicles, easy-to-access wireless channels, insufficient authentication, and inadequate trust among nodes are key problems in maintaining security and privacy within a network. In this survey, we first classified attacks over vehicular networks into four different groups that include hardware/software, infrastructure, sensors, and wireless communication. We then discussed six major requirements of security in vehicular networks where ML has been widely adopted to satisfy the requirements. Next, we presented a classification of ML approaches in the context of vehicular network security. We briefly explained the working mechanism of each approach. The ML techniques proposed in the literature for security of vehicular networks were described and summarized in tables for a clear understanding. While ML techniques bring in several benefits, they have different limitations which pose new challenges. We discussed such challenges in ML-based vehicular security that require further study. To provide useful insights, we presented our observations and lessons learned from this survey.

\balance

\bibliographystyle{IEEEtran}
\bibliography{IEEEabrv,References}

\end{document}